\def\paperlanguage{}
\pdfoutput=1 

\documentclass[letterpaper, 10 pt, conference]{ieeeconf}  

\usepackage{bm}
\usepackage{cite}
\include{preamble}

\IEEEoverridecommandlockouts                              

\title{\LARGE \textbf
  {
    \switchlanguage%
    {%
      Antagonist Inhibition Control in Redundant Tendon-driven Structures\\
      Based on Human Reciprocal Innervation\\
      for Wide Range Limb Motion of Musculoskeletal Humanoids
    }%
    {%
      人体の相反性神経支配に基づいた拮抗筋抑制制御による筋骨格腱駆動ヒューマノイドの上肢広可動域動作の実現
    }%
  }
}

\author{Kento Kawaharazuka, Masaya Kawamura, Shogo Makino, Yuki Asano, Kei Okada and Masayuki Inaba
  \thanks{Authors are with Department of Mechano-Informatics, Graduate School of Information Science and Technology, The University of Tokyo, 7-3-1 Hongo, Bunkyo-ku, Tokyo, 113-8656, Japan.
    {\texttt\small [kawaharazuka, kawamura, makino, asano, k-okada, inaba]@jsk.t.u-tokyo.ac.jp}
  }
}
\begin{document}

\maketitle
\thispagestyle{empty}
\pagestyle{empty}

\begin{abstract}
  \switchlanguage%
  {%
    The body structure of an anatomically correct tendon-driven musculoskeletal humanoid is complex, and the difference between its geometric model and the actual robot is very large because expressing the complex routes of tendon wires in a geometric model is very difficult.
    If we move a tendon-driven musculoskeletal humanoid by the tendon wire lengths of the geometric model, unintended muscle tension and slack will emerge.
    In some cases, this can lead to the wreckage of the actual robot.
    To solve this problem, we focused on reciprocal innervation in the human nervous system, and then implemented antagonist inhibition control (AIC) based on the reflex.
    This control makes it possible to avoid unnecessary internal muscle tension and slack of tendon wires caused by model error, and to perform wide range motion safely for a long time.
    To verify its effectiveness, we applied AIC to the upper limb of the tendon-driven musculoskeletal humanoid, Kengoro, and succeeded in dangling for 14 minutes and doing pull-ups.

  }%
  {%
    人体模倣型の筋骨格腱駆動ヒューマノイドの体の作りは非常に複雑であり、実機とモデルの間に大きな差異が生じる。
    なぜなら、複雑な筋の取り回しを完全に幾何モデルに落としこむことが非常に困難だからである。
    そして、幾何モデル通りに筋長指令を送れば、当然幾何モデルとの誤差により、意図しない筋張力や弛みの発生が伴う。
    これは時に実機の破損につながってしまう。
    我々はこの問題に対して、人体の反射である相反性神経支配に着目し、この考え方を模した拮抗筋抑制制御を開発し上肢に適用した。
    これによって、モデル誤差による内力の高まりや筋の弛みを回避しつつ、安全に長時間広可動域を行うことができる。
    検証として、肩、肩甲骨、肩と肩甲骨を使った腕上げ動作を行い、最後に14分間に及ぶぶら下がりと懸垂動作に成功した。
  }%
\end{abstract}

\section{INTRODUCTION} \label{sec:1}
\begin{figure}[htb]
  \centering
  \includegraphics[width=0.9\columnwidth]{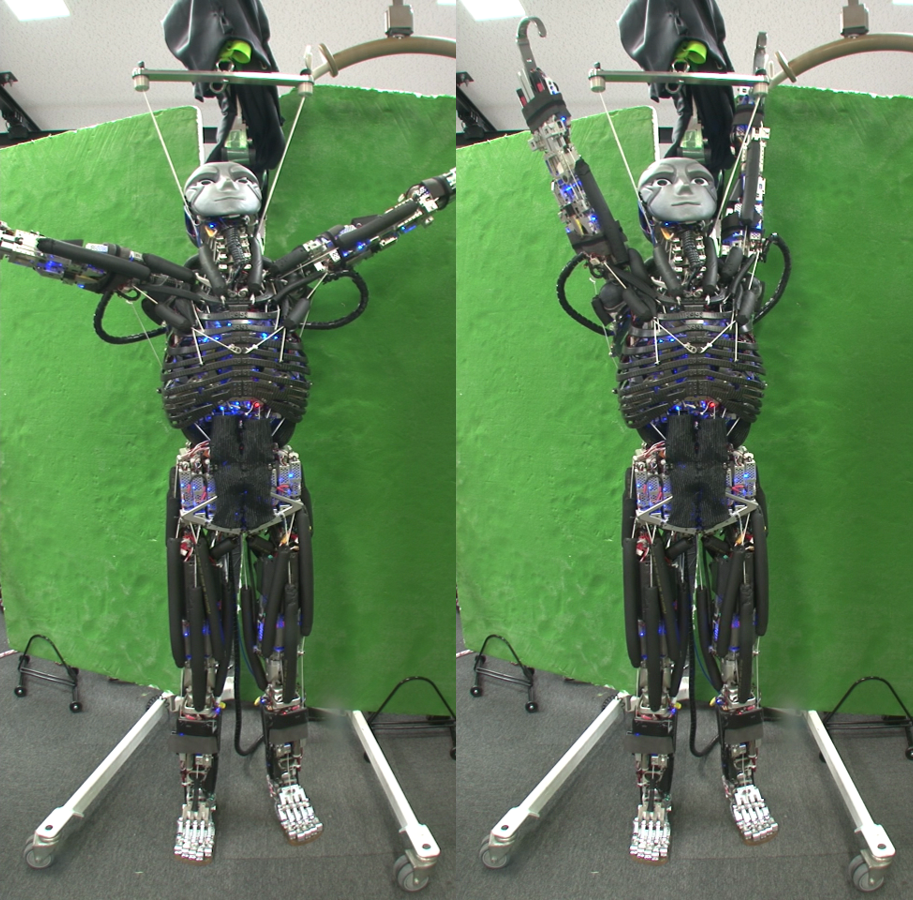}
  \caption{Example of wide range upper limb motion.}
  \label{figure:burasagari-overview}
\end{figure}

\switchlanguage%
{%
  In recent years, the development of tendon-driven musculoskeletal humanoids is very vigorous \cite{humanoids2013:michael:anthrob, humanoids2016:asano:kengoro}.
  So far, many topics such as its joint structure, arrangement of muscles, design approach and so on have been discussed.
  However, its joint structure and drive system are very complex because they are based on the human body.
  This becomes a big problem when moving the actual robot.
  For an ordinary robot, we make a geometric model and test the motion in a simulation environment.
  After that, we verify that the actual robot moves correctly.
  However, the difference between the geometric model and the actual robot in a tendon-driven musculoskeletal humanoid is very large because it is impossible to fully express the complex routes of tendon wires in a geometric model.

  To solve this problem, we can make a more complex and detailed geometric model, or implement a new control system which can absorb model error.
  There are some studies about modeling detailed tendon-driven musculoskeletal humanoids, but they are difficult to introduce, computationally complex, and unable to solve model error as much as an ordinary robot.
  Also, there are only a few studies moving the actual robot using the detailed model, much less doing wide range motion such as raising the arms using the scapula and shoulder (\figref{figure:burasagari-overview}), in which model error is fatal.
}%
{%
  近年、筋骨格腱駆動ヒューマノイドの開発は非常に盛んである\cite{humanoids2010:hugo:eccerobot}\cite{humanoids2016:asano:kengoro}。
  関節構造や筋配置、設計手法など、多くの議論が成されてきた。
  しかし、これらのロボットは人体を模倣しているがゆえに非常に複雑な関節構造、駆動方式をしている。
  それは実機を動かす際に大きな問題となる。
  通常の軸関節型ヒューマノイドであれば、幾何モデルを作成し、シミュレーション上でそのモデルの動作をテストする。
  その後、正しく動くことを実機を用いて検証する。
  しかし筋骨格腱駆動ヒューマノイドでは幾何モデルと実機の間に大きな差異が生じる。
  なぜなら、複雑な筋の取り回しを完全にモデル化することは不可能だからである。
  これを解決する方法としては、より複雑で詳細なモデル化の方向性と、モデルと実機の誤差を制御で解決する方向性が存在する。
  筋骨格腱駆動ロボットのモデル化に関していくつかの研究が存在するが、それらは非常に複雑であり導入や時間制約が難しいうえ、軸駆動ほど実機モデル誤差を解消することはできない。
  また、それらモデル化によって実機を動かした研究は少なく、ましてや、\figref{figure:burasagari-overview}のように誤差の効いてくる肩甲骨と肩を用いて腕を真上まで挙げるような広可動域動作を実現した研究は皆無である。
}%
\switchlanguage%
{%
  There are also some studies about the control of tendon-driven musculoskeletal humanoids.
  They include the simple control of movements by the wire length of the geometric model, muscle stiffness control \cite{robio2011:shirai:control}, position and force control with rectifiers \cite{csm1990:jacobsen:control}, puller-follower control \cite{ijars:potkonjak:puller-follower}, and joint-space control \cite{robio2011:michael:control, humanoids2016:kawamura:controll}.
  However, if we attempt to move the actual robot in a wide range using these controls, large internal muscle tension emerges due to model error, or the robot cannot move to the desired position  because it is assumed that the geometric model is correct.
  Therefore, there are few  experiments using not the geometric model but the actual tendon-driven musculoskeletal humanoid, much less one that achieves wide range motion.
  Also, experiments of the actual robot are almost always done using the uniaxial elbow joint with an encoder, and experiments using the ball joints in the shoulder or the complex scapula are done  by using motion capture, which needs large-scale setup.
  This is a big problem, and we believe that it interferes with the popularization of the tendon-driven musculoskeletal humanoid, which is the ultimate humanoid based on the human body.

  To move the tendon-driven musculoskeletal humanoid smoothly without accumulating internal muscle tension, we focused on the human nervous system.
  There are some reflexes in the human nervous system such as the stretch reflex, the tendon reflex, and reciprocal innervation, and among them, humans are able to move their bodies smoothly by reciprocal innervation, which inhibits antagonist muscles.
  We hypothesized that wide range motion could be achieved by applying this reciprocal innervation to the tendon-driven musculoskeletal humanoid.
  Although the control of pneumatic artificial muscles (PAMs) is similar to the control in this study in the sense that they both inhibit antagonist muscles, they are actually completely different.
  The control of PAMs controls muscles having simple antagonistic relationships by pressure, but the control in this study does position control according to the complex relationship of agonist and antagonist muscles as the antagonistic states correspond to the limb posture.
  There is the related work of muscle load sharing among agonist muscles \cite{humanoids2013:asano:loadsharing}, but what we want to focus on is the study between agonist muscles and antagonist muscles.
  Additionally, we aimed to develop a system in which an encoder or motion capture is unnecessary by using the estimation method of the joint angle \cite{humanoids2015:okubo:muscle-learning}.
  This method cannot estimate the joint angle of the complex shoulder or scapula very accurately, but the error is tolerated by using the joint angle only for the decision of whether a muscle is an agonist or antagonist muscle.

  In this study, we show that the tendon-driven musculoskeletal humanoid, which could not move in a wide range owing to large model error, can achieve wide range motion without accumulating internal muscle tension using the simple antagonist inhibition control (AIC) system based on reciprocal innervation.
  Then, we discuss the difference between this study and other control systems.
  %
}%
{%
  同様に、筋骨格腱駆動ロボットの制御に関する研究もいくつか存在する。
  単純にモデルの筋長を与えるものから、筋剛性制御\cite{robio2011:shirai:control}、position and force control with rectifiers\cite{csm1990:jacobsen:control}, puller-follower制御\cite{ijars:potkonjak:puller-follower}、Joint-Space Control\cite{humanoids2016:kawamura:controll}などが存在する。
  しかし、そのどれも幾何モデルの正しさを前提にしているがゆえ、広可動域を動作させようとすると実機との誤差によって筋張力が高まってしまうか、到達したい位置まで到達することができない。
  それゆえシミュレーションではなく実機で筋骨格腱駆動ヒューマノイドの動作実験を行ったものは少なく、ましてや、広く可動域を使う動作例はほとんどない。
  また、実機の動作例で制御に関節角度を用いる場合は、エンコーダを用いているため一軸の肘関節のみの実験が多く、球関節である肩や複雑な肩甲骨を用いた動作例の場合は大掛かりなセットの必要なモーションキャプチャを用いた実験がほとんどである。
  これは非常に問題であり、筋骨格腱駆動ヒューマノイドという、人間を最大限まで模した身体構造の究極系の普及に大きな支障をきたすと考える。

  そこで、人体構造を模した筋骨格腱駆動ヒューマノイドを内力を溜めずにスムーズに動作させるために、我々は人体の反射に着目した。
  人体の反射には伸張反射や腱反射、相反性神経支配などが存在するが、その中でも、相反性神経支配という主動筋に対して拮抗筋を抑制する機構によって人は体を柔軟に動かしている。
  この拮抗筋抑制の考え方をうまくヒューマノイドに適用することで、広可動域動作を実現できると考えた。
  Mckibben型アクチュエータの制御も拮抗筋を抑制するという意味では似ているが、姿勢によって拮抗関係の変わっていくなかで筋同士の関係から主動筋拮抗筋を分け位置制御をするという本研究の制御とは全く異なるものである。
  関連研究としては\cite{humanoids2013:asano:loadsharing}があるが、これは主動筋間に関する負荷分散の議論であり、我々が目指すところは主動筋拮抗筋間に関する議論である。
  また、\cite{humanoids2015:okubo:muscle-learning}の関節角度推定を用いることでエンコーダやモーションキャプチャを用いないシステムを目指した。
  この推定手法は複雑な肩関節や肩甲骨に関して良い精度は出ないものの、この関節角度推定を拮抗関係の判定のみに使うことによってその誤差を許容している。

  本研究では、実機とモデルの誤差によって内力が溜まり大きな動作の出来なかった筋骨格腱駆動ヒューマノイドの上肢において、人体の相反性神経支配を模した拮抗筋抑制制御を適用するによって、非常に単純なシステムで内力を溜めずに広可動域動作を実現できることを示す。
  そして、この制御法を他の制御方式と比較し、いくつかの議論を行う。

  \secref{sec:1}では本論文の動機と目標について述べた。
  \secref{sec:2}では本研究で使用する筋骨格腱駆動ヒューマノイド「腱悟郎」のワイヤ駆動方法や上肢の構造、また、幾何モデルと実機の誤差について説明を行う。
  \secref{sec:3}では人体の相反性神経支配を模して新しく開発した拮抗筋抑制制御のシステムについて述べる。
  \secref{sec:4}では拮抗筋抑制制御と誤差の関係、他の制御との比較を行う。
  \secref{sec:5}では拮抗筋抑制制御を適用した肩単体の動作、肩甲骨単体の動作、それらを統合して腕を真上まで挙げる動作を行い、他の制御との比較を行う。
  また、懸垂動作を行いこの制御法の利点と課題を明らかにする。
  最後に、\secref{sec:6}では結論と今後の方針について述べる。
}%

\section{System of the Tendon-Driven Musculoskeletal Humanoid, Kengoro} \label{sec:2}
\subsection{Tendon-Driven System of Kengoro}
\begin{figure}[htb]
  \centering
  \includegraphics[width=1.0\columnwidth]{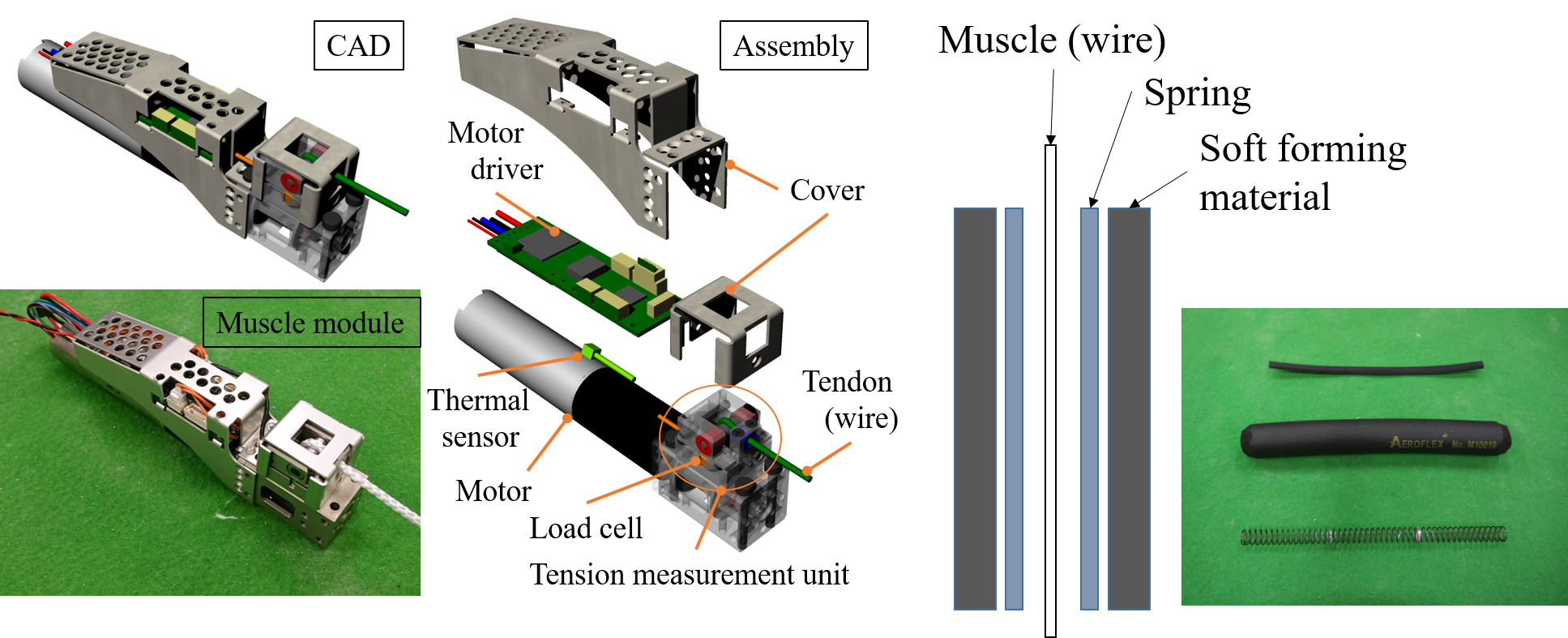}
  \vspace{-3.0ex}
  \caption{Structure of Kengoro's muscle module \cite{humanoids2016:asano:kengoro, iros2015:asano:module}. Left: sensor-driver integrated muscle module. Right: muscle wire and foam cover.}
  \label{figure:kengoro-muscle-structure}
\end{figure}

\switchlanguage%
{%
  The robot we use for this study is the tendon-driven musculoskeletal humanoid, ``Kengoro'' \cite{humanoids2016:asano:kengoro}.
  This robot is modeled after the human body, including its joint structure, body proportion, weight ratio, and joint performance, and is used as the research platform for the human body simulator and full-body contact behavior.
  The drive system is composed of 116 sensor-driver integrated muscle modules \cite{iros2015:asano:module} and duplicates the major muscles of a human.
  As \figref{figure:kengoro-muscle-structure} indicates, the muscle module has a brushless DC motor, a gear head, a pulley that winds wire, a Dyneema that acts as muscle wire, a temperature sensor that monitors muscle temperature, a tension measurement unit (weight limit: about 55 [kgf]), and a motor driver that can do current control.
  Dyneema is a chemical fiber that is resistant to abrasion.
  It can be a cause of model error because it extends by muscle tension as shown in \cite{humanoids2013:asano:loadsharing}.
  To allow flexible contact with the environment, there is a foam cover and a spring that prevents the inhibition of muscle movement around the Dyneema.
  However, this becomes a cause of difficulty when modeling the route of muscles.
  As reference, the gear ratio of this muscle module is 29:1 in most cases, and the temperature starts to rise when the muscle tension is more than about 30 [kgf].
}%
{%
  本研究で用いるロボットは筋骨格腱駆動ヒューマノイド「腱悟郎」\cite{humanoids2016:asano:kengoro}である。
  人体における関節構造・身体プロポーションや重量・筋配置・関節性能を模しており、人体シミュレータとしての活用や全身環境接触行動のための研究用プラットフォームとして活躍している。
  腱駆動方式は\cite{iros2015:asano:module}の筋モジュールが全身に116個配置され、人間の主要な筋肉を模している。
  筋モジュールには\figref{figure:kengoro-muscle-structure}のようにブラシレスDCモータとギアヘッド、糸を巻き取るためのプーリと筋としてのDyneema、筋の温度を監視するための温度センサ、筋張力を測定するための張力測定ユニット(定格約55[kgf])、電流制御可能なMotor Driverが格納されている。
  Dyneemaは摩擦に強い化学繊維であり、\cite{humanoids2013:asano:loadsharing}に示されるように力によって多少の伸びを生じ、誤差の原因と成り得る。
  また、筋としてのDyneemaの周りには、発泡性の柔らかいカバーと筋の動きを阻害しないためのバネがついており、これによってより柔軟な環境への接触を実現しているが、同時に筋経路のモデル化が難しい一つの要因となっている。
  参考として、ギア比は基本的に29:1であり、おおよそ30[kgf]以上の力で発熱をする。
}%

\subsection{Structure of Kengoro Upper Limb}
\begin{figure}[htb]
  \centering
  \includegraphics[width=1.0\columnwidth]{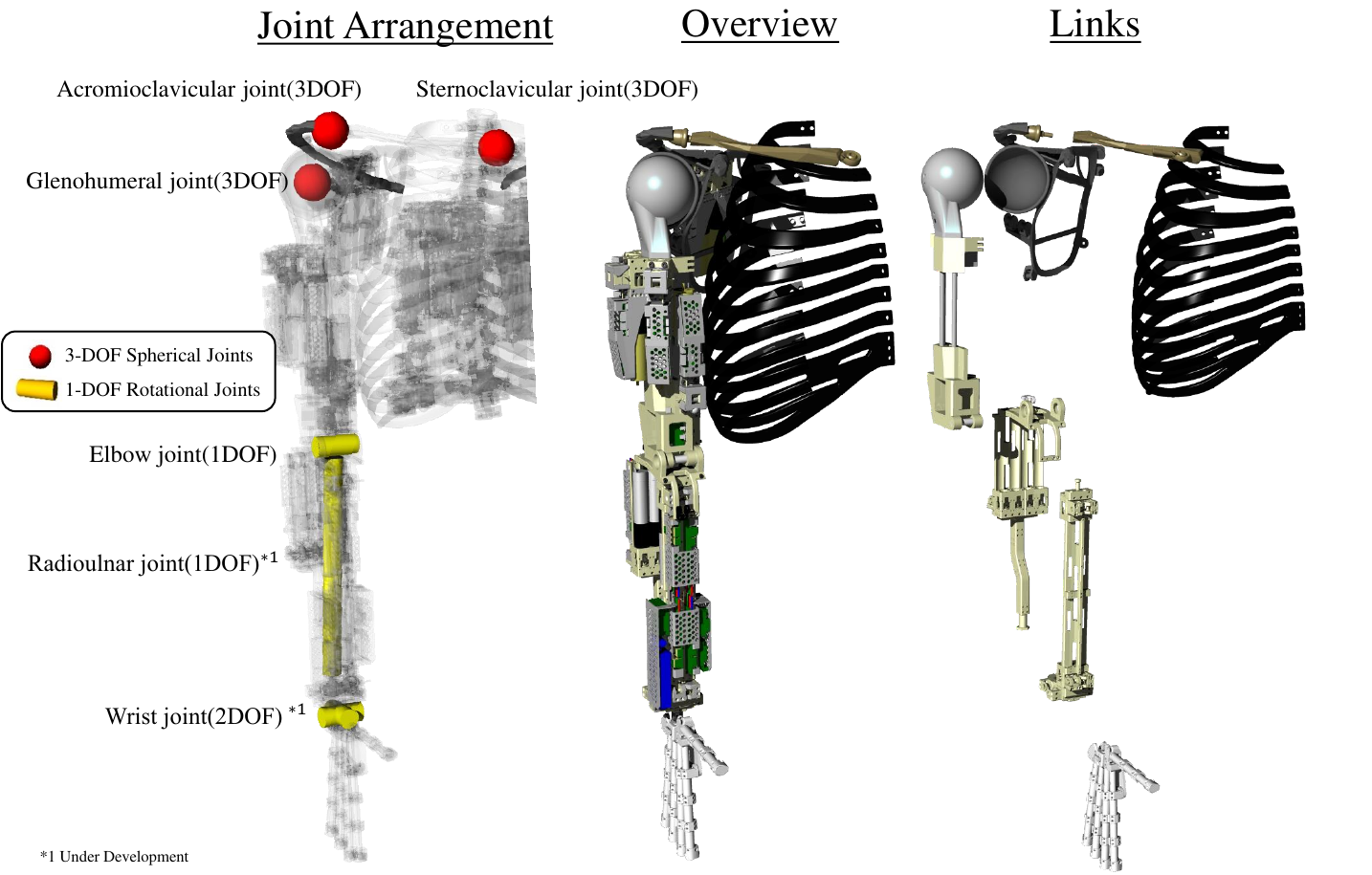}
  \vspace{-3.0ex}
  \caption{Structure of Kengoro upper limb.}
  \label{figure:kengoro-upper-limb-structure}
\end{figure}

\switchlanguage%
{%
  The human upper limb extends from the sternum to the clavicle, scapula, humerus, ulna, and radius, in order.
  The sternum and clavicle are connected by the sternoclavicular (SC) joint, the clavicle and scapula are connected by the acromioclavicular (AC) joint, the scapula and thorax are connected by the scapulothoracic (ST) joint, and the scapula and humerus are connected by the glenohumeral joint (GH).
  This structure of the human body applies to Kengoro, as shown in \figref{figure:kengoro-upper-limb-structure}.
  The movement of the shoulder is composed of the GH joint, which is a ball joint that has 3 DOFs.
  The movement of the scapula is composed of a 3 DOFs joint (upward rotation, downward rotation, adduction, abduction, elevation, depression) because the SC and AC joints, having 6 DOFs total, are constrained by the ST joint.
  We cannot raise our arms with just the GH joint in the shoulder.
  We are able to raise our arms by the upward rotation of the scapula, along with the movement of the shoulder joint.
  This ratio is said to be 2:1, and that the scapula rotates upward by 60 [deg] when the abduction of the shoulder is by 120 [deg].
  Therefore, simultaneous movement of the scapula and shoulder is important for wide range upper limb motion such as touching one's back and dangling.
}%
{%
  人体の上肢は胸骨から伸びており、鎖骨、肩甲骨、上腕、前腕の順で骨が繋がっている。
  胸骨と鎖骨が胸鎖関節によって、鎖骨と肩甲骨が肩鎖関節によって、肩甲骨と上腕が肩甲上腕関節によって繋がっている。
  これは腱悟郎でも全く同様であり、\figref{figure:kengoro-upper-limb-structure}のような構造となっている。
  肩の肩甲上腕関節は球関節で3自由度であり、肩甲骨は胸鎖関節と肩鎖関節の6自由度を胸郭と肩甲骨の接触によって拘束し、3自由度(上方回旋下方回旋、内転外転、挙上下制)を形成している。
  人間の手を真上まで挙げるような動作の際、肩の3自由度だけではそれを成すことはできない。
  人間は肩の動作と同時に、肩甲骨を上方回旋させることによってそれを成している。
  その比は2:1と言われ、肩がroll方向に120度動く間に肩甲骨は60度上方回旋をすると言われる。
  よって、人間が自分の背中を触ったり、棒にぶら下がったりするような広く可動域を使った動作をするためには、この肩甲骨と肩を連動させて動作させることが重要となる。
}%

\subsection{Comparison Between Geometric Model of Kengoro and Actual Kengoro}

\begin{figure}[htb]
  \centering
  \includegraphics[width=0.9\columnwidth]{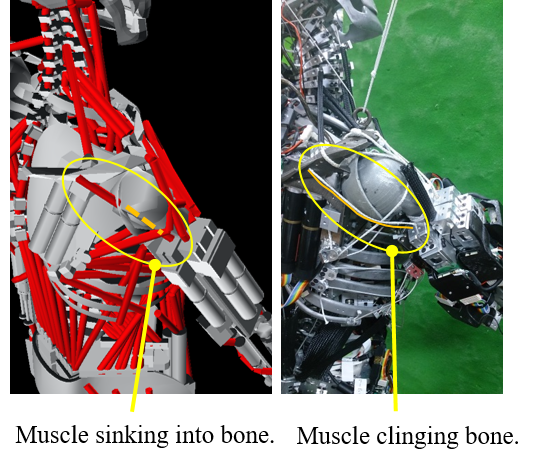}
  \caption{Geometric model and actual robot.}
  \label{figure:what-model-error-is}
\end{figure}

\switchlanguage%
{%
  The geometric model of the tendon-driven musculoskeletal humanoid is very complex and difficult.
  In the geometric model of Kengoro, we model the route of muscle wires by arranging the start point, end point, and multiple relay points.
  However, this method has a big problem.
  The problem is that we cannot model the route of the wires perfectly because, as shown in \figref{figure:what-model-error-is}, the muscle of the actual robot clings around the structure, but the muscle of the geometric model sinks into the structure.
  At the same time, due to the complex structure, the route of muscle wires can deviate from the desired path depending on the joint angle.
}%
{%
  筋骨格腱駆動ヒューマノイドの幾何モデルは非常に複雑で難しい。
  腱悟郎の幾何モデルは筋肉の起始点と終止点、そして複数の中継点を設けることで筋の取り回しを模している。
  しかし、これには非常に大きな問題がつきまとう。
  それは、この方法では筋経路を完全にモデル化することはできず、\figref{figure:what-model-error-is}に示すように実機では筋が骨格にまとわりつくようになるのに対して、幾何モデルでは筋が骨格にめり込んでしまうことがあるという問題である。
  また、筋経路が関節角度によってズレてしまうことも複雑な構造を持つがゆえに起こりえるのである。
}%

\section{System of Antagonist Inhibition Control} \label{sec:3}
\subsection{Reciprocal Innervation in Humans}
\begin{figure}[htb]
  \centering
  \includegraphics[width=0.9\columnwidth]{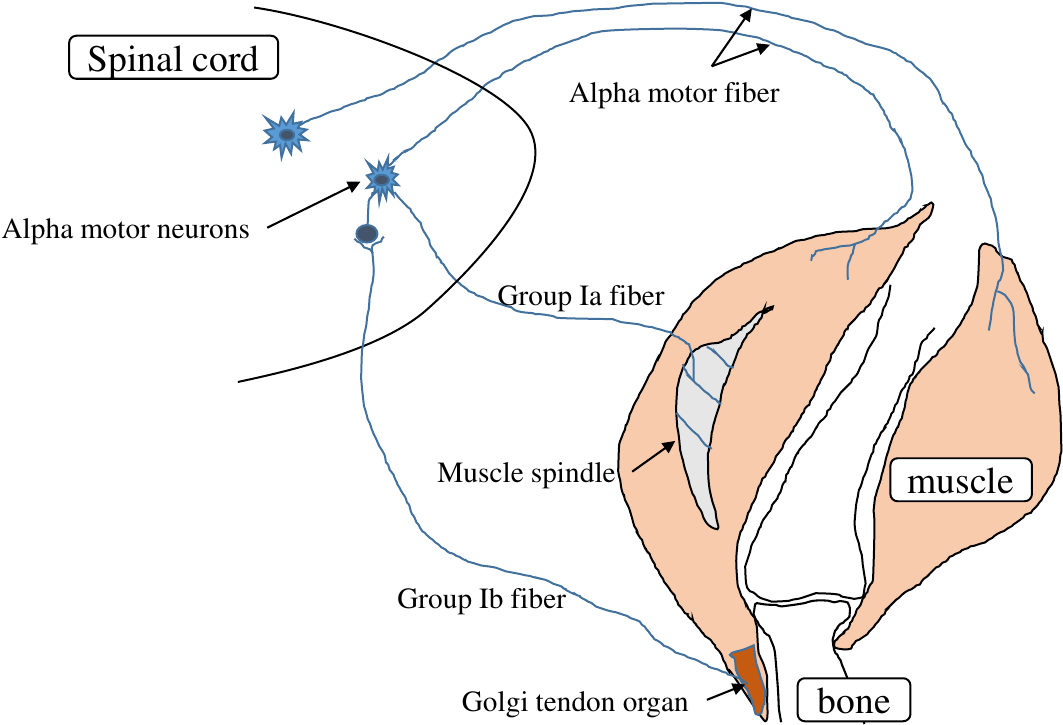}
  \caption{The principle of human reciprocal innervation.}
  \label{figure:human-antagonist-inhibition}
\end{figure}

\switchlanguage%
{%
  First, we want to consider how humans control their bodies.
  The principle of the human nervous system is indicated in \figref{figure:human-antagonist-inhibition}.
  There are sensory receptors called the muscle spindle and the tendon organ in the muscle.
}%
{%
  まず、人間がどのように体をコントロールしているかについて考える。
  人間の反射系は\figref{figure:human-antagonist-inhibition}に示すような構造になっている。
  筋肉には筋紡錘・腱器官という感覚受容器が備わっている。
}%

\switchlanguage%
{%
  The muscle spindle is a spindle-shaped organ adhered parallel to the muscle fiber, and is a receptor that detects muscle length and shrinkage velocity.
  This creates a reflex loop called the stretch reflex, and consists of a negative feedback system based on muscle length.

  The tendon organ is arranged at the edge of the muscle, and is connected to the muscle fiber in series.
  This creates a reflex loop called the tendon reflex, which consists of a negative feedback system based on muscle tension.

  What we want to focus on in this study is reciprocal innervation in the human nervous system.
  Muscles can create tension only in the direction of contraction, so one or more pairs of antagonistic muscles have to exist in order for the smooth movement of the joints to occur.
  Of the antagonistic muscle pairs, muscles that contract in the direction of joint movement are called agonist muscles, and the others are called antagonist muscles.
  Therefore, to move the body smoothly, we need to stimulate $\alpha$ motor neurons of agonist muscles and inhibit those of antagonist muscles.
  In the human body, a neural circuit that stimulates motor neurons of agonist muscles and at the same time inhibits those of antagonist muscles is equipped.
  Such reciprocal interaction between agonist and antagonist muscles is called reciprocal innervation.
  This neural circuit between muscles is complex, and in this study, we developed antagonist inhibition control by imitating this complex reciprocal action.
}%
{%
  筋紡錘は筋繊維に平行に付着する紡錘形の器官であり、筋の長さ・短縮速度を検知する受容器である。
  これは伸張反射と呼ばれる反射ループを成し、筋長を出力とする負のフィードバック系を成す。

  腱器官は筋の端に配置されており、筋紡錘が筋繊維と平行に配置されているのに対し、筋繊維と直列に接続する。
  これは腱反射系とよばれる反射ループを成し、筋張力を出力とする負のフィードバック系を成す。

  そして、本研究で焦点を当てるのは、相反性神経支配という神経回路である。
  筋肉は収縮する方向のみに張力を発生できるが、それゆえ自由に関節を動作させるために人体には関節の両側に拮抗的に働く一対以上の筋が存在する。
  このとき拮抗関係にある筋において、関節を動作させる方に働く筋を主動筋、その他を拮抗筋と呼ぶ。
  つまり、スムーズに体を動作させるためには主動筋の運動ニューロンを興奮させ、拮抗筋の運動ニューロンを抑制する必要がある。
  ここで、筋紡錘からのIa求心性繊維は主動筋の運動ニューロンを興奮させると同時に、抑制性のニューロンを介して拮抗筋の運動ニューロンを抑制するような神経回路が用意されている。
  このような拮抗筋と主動筋の運動ニューロン間の相互作用を相反性神経支配という。
  この神経回路は筋肉間で複雑な神経系を成しており、本研究ではこの筋肉同士の複雑な相互作用を擬似的に模すことで拮抗筋抑制制御を開発した。
}%

\begin{figure}[htb]
  \centering
  \includegraphics[width=1.0\columnwidth]{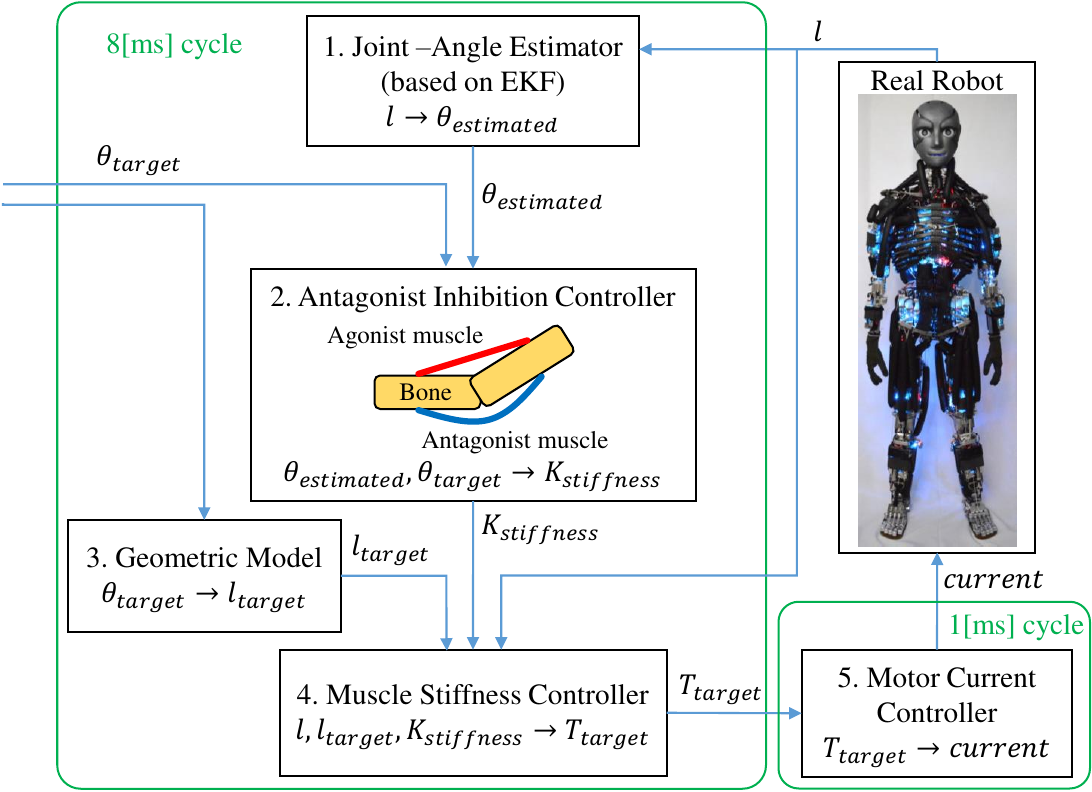}
  \vspace{-3.0ex}
  \caption{System of antagonist inhibition control.}
  \label{figure:kengoro-antagonist-inhibition}
\end{figure}

\subsection{System of Antagonist Inhibition Control}
\switchlanguage%
{%
  The newly implemented antagonist inhibition (AIC) system is a simple control system to change the muscle stiffness gain of muscle stiffness control (MSC) between agonist and antagonist muscles.
  The overview of the system is shown in \figref{figure:kengoro-antagonist-inhibition}, and each topic will be explained respectively.

  \begin{enumerate}
    \item \textbf{Joint-Angle Estimator}

      We must estimate the joint angle from the change in muscle length because the angle of the ball joint cannot be measured by the rotary encoder, potentiometer, and so on.
      We use the estimation method of the joint angle proposed by Okubo, et al. \cite{humanoids2015:okubo:muscle-learning}.
      We can estimate the joint angle of the actual robot from the change in muscle length by Extended Kalman Filter.

    \item \textbf{Antagonist Inhibition Controller}

      This is the most important part in this study.
      We want to inhibit muscle tension of antagonist muscles like in reciprocal innervation.
      Muscle jacobian $G(\bm{\theta})$ is expressed as below.
      \begin{align}
        G(\bm{\theta}) = d\bm{l}/d\bm{\theta}
      \end{align}
      This $(l{\times}n)$ matrix ($l$ is the number of muscles, $n$ is the number of joints) expresses how much the muscle contracts when the joint moves in a certain direction while the joint angle is $\bm{\theta}$.
      In other words, when the target joint angle is $\bm{\theta}_{target}$, and the current joint angle is $\bm{\theta}$, $G(\bm{\theta})(\bm{\theta}_{target}-\bm{\theta})$ expresses whether each muscle is an agonist muscle or an antagonist muscle when moving to that position.
      In this study, we change the muscle stiffness $K_{stiffness}$ depending on if the muscle is an agonist or antagonist muscle.
      When $K_{stiffness}$ is positive, the muscle causes tension in the direction of the target length $\bm{l}_{target}$, and when $K_{stiffness}$ is 0, the muscle tension is constant at $\bm{T}_{bias}$.
      These represent agonist and antagonist muscles exactly, and the AIC system decides as below regarding the $i$-th muscle:
      \begin{align}
        s = G(\bm{\theta})\frac{\bm{\theta}_{target}-\bm{\theta}}{|\bm{\theta}_{target}-\bm{\theta}|}[i] \\
        \bm{K}_{stiffness}[i] = k \;\;\;\;\;\; if \;\; s < C \\
        \bm{K}_{stiffness}[i] = 0 \;\;\;\;\;\; if \;\; s \geqq C
      \end{align}
      where $s$ represents moment arm in the intended direction by the normalization of $(\bm{\theta}_{target}-\bm{\theta})$, $k$ is the constant value given to $\bm{K}_{stiffness}$ of agonist muscles, and $C$ is the threshold value which decides the antagonistic state.
      To stabilize movement, we change $\bm{K}_{stiffness}$ linearly from $0$ to $k$, and then from $k$ to $0$ in $t_k$[msec].

      Also, we will show how to obtain the muscle jacobian.
      There are polyarticular muscles in the human body; for example, the pectoralis major muscle is influenced by the movement of the scapula and shoulder joints.
      Like the method proposed by Okubo, et al. \cite{humanoids2015:okubo:muscle-learning}, we focus on a certain muscle, and formulate its muscle length as a polynomial of related joint angles using data from the geometric model.
      Then, we differentiate the polynomial to obtain the muscle jacobian.

    \item \textbf{Geometric Model}

      When we give the target joint angle to the geometric model, such as in the left picture of \figref{figure:what-model-error-is}, we can find the target muscle length from the start point, the relay points, and the end point of the muscle.

    \item \textbf{Muscle Stiffness Controller}

      In MSC, the target muscle tension is decided by multiplying muscle stiffness by the difference between the target muscle length and the current muscle length.
      \begin{align}
        \bm{T}_{target} = \bm{T}_{bias} + \textrm{max}\{0, \bm{K}_{stiffness}(\bm{l}-\bm{l}_{target})\}
      \end{align}
      The $\bm{T}_{bias}$ is added in order to prevent the loosening of the muscle.

    \item \textbf{Motor Current Controller}

      This is the part that converts from the target muscle tension to the actual current that flows to the muscle motor.

  \end{enumerate}
}%
{%
  本研究で開発した拮抗筋抑制制御について説明する。
  概要としては、筋剛性制御の剛性係数を、主動筋と拮抗筋で変えるというシンプルな制御である。
  システム構成は\figref{figure:kengoro-antagonist-inhibition}のようになっており、一つ一つの要素を順に説明する。
  \begin{enumerate}

    \item Joint-Angle Estimator

      これはロボットの関節角度を推定する部分である。
      球関節はエンコーダ等によって角度を知ることができないため、筋長変化から関節角度を推定しなければならない。
      推定方法は大久保らの方法\cite{humanoids2015:okubo:muscle-learning}を用いる。
      実際のロボットの筋長変化からExtended Kalman Filterを用いて現在のロボットの関節角度を推定している。

    \item Antagonist Inhibition Controller

      本研究で最も重要なControllerである。
      人間の相反性神経支配のように、拮抗筋の力を抑制することを考える。
      筋長ヤコビアン$G(\bm{\theta})$は
      \begin{align}
        G(\bm{\theta}) = d\bm{l}/d\bm{\theta}
      \end{align}
      と表されるように、関節角度が$\bm{\theta}$の際に、ある方向に関節を動かした際にどれだけ筋が縮むかを表している$(l{\times}n)$の行列である($l$は筋数、$n$は関節数)。
      つまり、動かしたい方向を$\bm{\theta}_{target}$、現在の関節角度を$\bm{\theta}$とすれば、$G(\bm{\theta})(\bm{\theta}_{target}-\bm{\theta})$はそれぞれの筋がその方向に動くときに主動筋となるか、拮抗筋となるかを表している。
      本研究では筋剛性制御の筋剛性$K_{stiffness}$を主動筋か拮抗筋かによって変更する。
      $K_{stiffness}$は0であればその筋の張力は$\bm{T}_{bias}$で一定となり、$K_{stiffness}$が正であればターゲットとなる筋長$\bm{l}_{target}$に追従する方向に力を発生させる。
      よってこれはまさに拮抗筋と主動筋であり、Antagonist Inhibition Controllerは以下のような決定をすることになる。
      i番目の筋に関して、
      \begin{align}
        s = G(\bm{\theta})\frac{\bm{\theta}_{target}-\bm{\theta}}{|\bm{\theta}_{target}-\bm{\theta}|}[i] \\
        \bm{K}_{stiffness}[i] = k && if && s < C \\
        \bm{K}_{stiffness}[i] = 0 && if && s \geqq C
      \end{align}
      となる(ここで$s$は$\bm{\theta}_{target}-\bm{\theta}$で割ることでモーメントアームを表すようになり、$k$は主動筋の$\bm{K}_{stiffness}$に与えるある定数, $C$は拮抗関係を決める閾値である)。
      動作を安定させるために、$\bm{K}_{stiffness}$の値は0からkに、そしてkから0に、$t_k$ミリ秒で線形に遷移させている。

      また、筋長ヤコビアンの求め方について示す。
      筋肉には多関節なものも含まれ、例えば大胸筋は肩甲骨や肩関節の動きに影響を受ける。
      そのため、Okuboらの手法と同じように、ある筋に着目しモデルのデータを用いて、その筋に関与する関節の角度の多項式として筋長を表す。
      それを微分し、muscle jaccobianを求めている。

    \item Kengoro Geometric-Model

      \figref{figure:what-model-error-is}の左図に示すような幾何モデルにターゲットとなる関節角度を与えることで、起始点、中継点、終始点からそれぞれの筋の指令値となる長さを知ることができる。

    \item Muscle Stiffness Controller

      ターゲットとなる筋の長さと現在の筋の長さの差に対して拮抗筋抑制制御で得た筋剛性をかけることでその張力とする。
      \begin{align}
        \bm{T}_{target} = \bm{T}_{bias} + \bm{K}_{stiffness}(\bm{l}-\bm{l}_{target})
      \end{align}
      式のように、筋が弛んでしまわないように$\bm{T}_{bias}$項を足しあわせている。

    \item Motor Current Controller

      これは張力目標値から、モータに流す実際の電流値に変換する部分である。

  \end{enumerate}
}%

\section{Characteristics of\\Antagonist Inhibition Control} \label{sec:4}
\begin{figure}[htb]
  \centering
  \includegraphics[width=0.9\columnwidth]{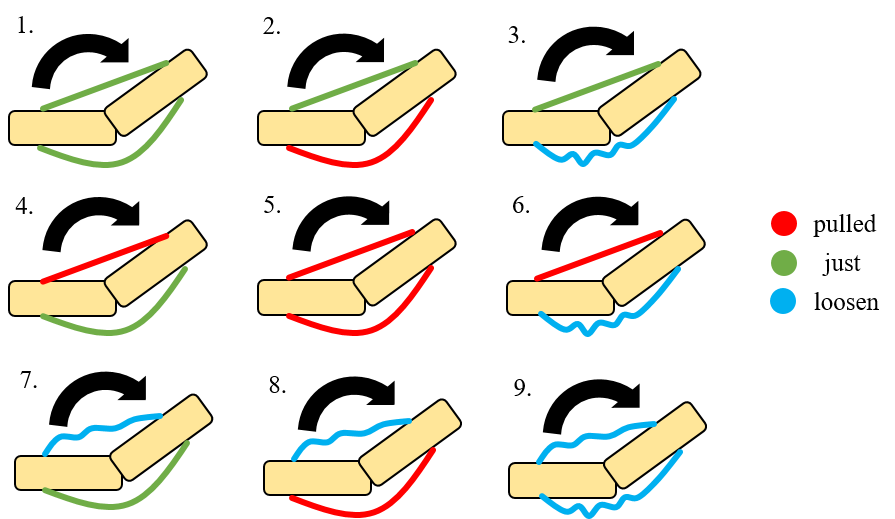}
  \caption{Various types of antagonistic model error.}
  \label{figure:various-model-error}
\end{figure}

\switchlanguage%
{%
  In this section, we will show some characteristics of AIC in order to better understand the benefits.

  \subsection{Analogy with Human Reciprocal Innervation}
  We mentioned above that human reciprocal innervation decides antagonistic states by the complex interaction between muscles.
  On the other hand, the operation flow of AIC is shown as below.
  First, in AIC, the joint angle is estimated with a complex formula using changes in muscle lengths \cite{humanoids2015:okubo:muscle-learning}.
  Then, the muscle jacobian of a certain muscle is obtained as a polynomial of the involved joint angles.
  AIC moves based on the antagonistic states obtained by the estimated joint angle, muscle jacobian, and intended posture.
  This means that AIC is based on the complex interactions of the changes in muscle lengths in the time direction, and that AIC can artificially imitate the complex muscle interactions of human reciprocal innervation.
  Also, the value $s$ obtained by the antagonistic states is the degree of excitation and inhibition of the $alpha$ motor neuron.
  AIC imitates human reciprocal innervation by regarding the excitation as the control that follows the target length, and inhibition as the control that keeps constant muscle tension.

  \subsection{Classification of Antagonistic States}
  There are 9 types of states in antagonistic muscles, shown as types 1--9 in \figref{figure:various-model-error}.
}%
{%
  本章では、拮抗筋抑制制御と相反性神経支配の類似性、実機とモデルとの誤差の種類の分類、拮抗筋抑制制御のもう一つの単純な制御との比較、拮抗筋抑制制御のパラメータなどの詳細の議論、基本実験をすることによって理解を深める。

  \subsection{Analogy with Human Reciprocal Innervation}
  相反性神経支配が多くの筋の間における複雑な相互作用であることは前に述べた。
  拮抗筋抑制制御は、この筋間相互作用を擬似的に模すことによって成り立っている。
  まず、拮抗筋抑制制御では関節角を推定する際には筋の長さ変化を用いることで複数の筋から複雑に関節角度が推定される。
  そして、ある筋に関しする筋長ヤコビアンは、関与する関節群の多項式として得られる。
  この関節角度と筋長ヤコビアン、動かしたい方向に対する意識から拮抗関係が推定され、それを元に動作するのが拮抗筋抑制制御である。
  これは擬似的に、多数の筋の複雑な時間方向に対する変化の絡み合いにより成り立っていることとなり、擬似的に相反性神経支配の筋間における複雑な相互作用を模していると言えるだろう。
  また、最終的にこれは筋長に関するRNNとして表せるだろう。

  \subsection{Classification of Antagonistic States}
  拮抗関係の状態には\figref{figure:various-model-error}の1〜9のように9つの誤差の種類が存在する。
}%

\switchlanguage%
{%
  There are 3 types of muscle tension states, which are just right, loose, and tight, and they result in 9 combinations of antagonistic states.
  When the joint moves in the direction of the arrow, the antagonistic state changes among these 9 states, and finally stops at states 10--13, which are stable.
  Among the final states, 10 is the best state because it has no model error and can keep adequate muscle tension.
  11 and 12 are states in which the muscle length of the actual robot is consistent with that of the geometric model, but either one could become loose due to model error, and 13 is the state in which the muscle wires pull each other, resulting in high muscle tension.
  In these states, the most dangerous state is 13.
  In state 13, the larger the model error is, the more the muscles pull each other, and can lead to fatal damage to the muscles or bones.
  The most important benefit of AIC is that it can avoid state 13 completely because it does not assume that its geometric model is absolutely correct.
  In AIC, antagonist muscles do not follow the target muscle length, and keep a constant muscle tension, so state 13 always stops at state 10.
  AIC can also prevent the slack of muscle wires using the same principle.
  On the other hand, if we assume that the geometric model is correct in spite of the model error, we cannot avoid states 11, 12, and 13.

  \subsection{Comparison Between AIC and Another Simple Control}
  We propose another new control and discuss its differences with AIC for better understanding.
  In AIC, we use whether the value as stated below is positive or negative in order to decide between agonist or antagonist muscle.
  \begin{align}
    s = G(\bm{\theta})(\bm{\theta}_{target}-\bm{\theta})[i]
  \end{align}
  where we have removed the denominator $|\bm{\theta}_{target}-\bm{\theta}|$ for easier understanding.
  The value as stated above seems to equal the value shown below because $\bm{\theta}_{target}-\bm{\theta}$ is $d\bm{\theta}$, and  $G(\bm{\theta})d\bm{\theta}$ is $d\bm{l}$.
  \begin{align}
    s = (\bm{l}_{target}-\bm{l})[i]
  \end{align}
  Thus, we can propose another control system using the value above for the decision of agonist or antagonist muscle.
  We will refer to the basic AIC as joint-based AIC, and the now proposed AIC as the muscle-based AIC.
  In joint-based AIC, when the muscles are antagonistic in the geometric model, the antagonist muscle is inhibited against the agonist muscle.
  This can avoid state 13 completely.
  On the other hand, in muscle-based AIC, muscles contract simply if they want to contract, and muscles keep constant muscle tension if they want to extend.
  In this case, the muscles could end up in state 13 because if the antagonist muscle contracted more than $\bm{l}_{target}$ due to model error, the antagonistic muscles pull each other.
  At a glance, the two controls seem to be the same, but muscle-based AIC is not practical because it cannot avoid state 13.
  The difference is that whereas muscle-based AIC uses the current muscle length as it is, joint-based AIC uses the estimated joint angle from the current muscle length.
  If we use the muscle length of the actual robot, the model error becomes large, but if we convert the actual muscle length to the joint angle of the geometric model, it is not influenced by model error because it has the idealistic conditions of the model.
  Additionally, we only decide between agonist or antagonist muscle in AIC, and this makes it hard for antagonistic model error to occur.
  For example, at the uniaxial joint shown in \figref{figure:various-model-error}, the model error due to muscle length can of course occur, but the antagonistic state does not change between the geometric model and the actual robot.

  \subsection{Other Characteristics}
  First, we will focus on the constant value $C$ used in the decision of antagonistic states.
  $C$ expresses the threshold of the moment arm in the intended direction.
  However small the moment arm of the muscle is, when $C$ equals 0, if $s$ is negative, the muscle is an agonist muscle, and vice versa.
  This is theoretically correct, so we set $C$ equal to 0 in this study.
  However, the muscle jacobian $G(\theta)$ is obtained by the geometric model, the moment arm of the muscle has model error in a certain degree, and the correct antagonistic states are not necessarily obtained by whether the moment arm is positive or negative.
  Although incorrect antagonistic states can emerge by model error, the muscles making incorrect antagonistic states have only small moment arm, so internal muscle tension does not accumulate.
  Also, if $C$ becomes bigger in the negative direction, all of the agonist muscles which have only small moment arm in the intended direction are inhibited, and the effect of antagonist inhibition becomes bigger.
  Thus, even if there is model error, we can make sure that incorrect antagonistic states do not emerge.
  Since movement in the unintended direction is permitted to a certain degree, we can see that the joint moves in the direction in which the muscle tension between agonist muscles become equal.
  However, at the same time, this creates unintended movement.
  On the other hand, if $C$ becomes bigger in the positive direction, the effect of antagonist inhibition becomes smaller, but this does not create unintended movement because the antagonist muscles that have small moment arm in the direction of movement inhibition act as agonist muscles.

  Next, we will focus on the estimated joint angle.
  The estimation result of a complex joint cannot be said to be very accurate.
  However, in AIC, agonist muscles of the actual robot follow the geometric model, and the estimated joint angle is only used for the decision of the antagonistic state.
  Thus, the inaccuracy of the estimated joint angle does not become a big problem because it is hard for antagonistic model error to occur.
}%
{%
  筋の状態には、ちょうどよい張力を保っているとき、緩んでいるとき、高張力となっているときの3種類が存在し、その組み合わせとしての9種類である。
  これらは矢印の方向に動作する際、この9つの状態を遷移しながら安定した状態である10〜13の状態に停止する。
  最終的な状態として、10は最も良い、実機とモデル誤差がないため緩むこともせず、引っ張り合うこともせず程よい張力を保った状態である。
  それに対して、11,12では両方ともモデルと完全に筋長が一致してはいるものの実機との誤差によって片方が緩んでしまうような状況であり、13は拮抗関係にある筋が引っ張り合い、高張力を発揮しあってしまうような状況である。
  この状態において、最も危険な状態は13の状態である。
  これは、実機とモデルの誤差が大きければ大きいほどより高張力で拮抗し合い、筋や骨の損傷となりかねない。
  そして拮抗筋抑制制御の最大の利点は、この13の状況を必ず回避できる点にあるのである。
  拮抗筋抑制制御においては拮抗筋は指令した筋長に追従せずに一定張力を発揮し続けるため、13は必ず10の状況に落ち着くことになる。
  また、拮抗筋は一定張力を発揮し続けるため、11,12の状況も10に落ち着くことになり、筋の弛みを防止することもできる。
  対して、筋骨格腱駆動ヒューマノイドの完全な幾何モデルを作るのは難しく、モデルの正しさを前提にしてしまった場合には11,12,13の状況を回避することができないのである。

  \subsection{Comparison Between AIC and Another Simple Control}
  一つ新しい制御を考案し、それと拮抗筋抑制制御の違いを論じてみる。
  拮抗筋抑制制御では、主動筋か拮抗筋かの判断に
  \begin{align}
    s = G(\bm{\theta})(\bm{\theta}_{target}-\bm{\theta})[i]
  \end{align}
  の正負を用いていた(簡単のため、分母の$\bm|{\theta}_{target}-\bm{\theta}|$は除いている)。
  この式において、$\bm{\theta}_{target}-\bm{\theta}$は$d\bm{\theta}$、$G(\bm{\theta})d\bm{\theta}$は$d\bm{l}$であるため、
  \begin{align}
    s = (\bm{l}_{target}-\bm{l})[i]
  \end{align}
  と同値であるように見え、この式を拮抗筋抑制制御における主動筋拮抗筋の判断基準に使う制御を考える。
  元々の拮抗筋抑制制御を関節based拮抗筋抑制制御、今考えた拮抗筋抑制制御を筋長based拮抗筋抑制制御と呼ぶこととする。
  関節basedでは、モデルの中で拮抗していれば必ず主動筋に対して拮抗筋が抑制されることになる。
  つまり、必ず13の状況を回避することができるのである。
  それに対して筋長basedでは、筋長を縮ませたい場合は単純に縮ませ、筋長を伸ばしたい場合は一定の張力をかけ続ける。
  この場合、13の状況に成り得るのである。
  なぜなら動作の際、拮抗筋がモデル誤差によって$\bm{l}_{target}$よりも縮んでしまった場合、主動筋と拮抗筋で引き合ってしまうからである。
  一見同じように見える２つの制御であるが、筋長basedな拮抗筋抑制制御は13の状況を回避できず役に立たない。
  この違いは、筋長basedではまさに現在の筋長をそのまま使っているのに対し、関節basedでは、現在の筋長から推定された関節角度を用いることにある。
  実機の筋長をそのまま使うとモデルと実機の誤差を大きく受けてしまうのに対して、実機の筋長を一度関節角に変換してモデルの中に落とし込んでしまえば、それは理想的なモデルの中であるため誤差の影響を受けないのである。
  そして、拮抗筋抑制制御では主動筋と拮抗筋の判定のみを行うが、この拮抗関係はモデル誤差が生じにくいのも大きな点である。
  例えば\figref{figure:various-model-error}のような一軸関節では、筋長に関してのモデル誤差は当然生じ得るが、拮抗関係はモデルであっても実機であっても変化することはないのである。

  \subsection{Other Characteristics}
  まず、拮抗関係の判定に用いられる定数$C$に関してである。
  $C$は動かしたい方向に対するモーメントアームの大きさの閾値を表している。
  Cが0の場合はモーメントアームがどれだけ小さかろうと0より小さいならば主動筋、0より大きいならば拮抗筋とすることを意味する。
  これは本来のあるべき姿なため、本研究では基本的にはC = 0と置いている。
  しかし、muscle jacobianはモデルから求めており、完璧にG(θ)(⊿θ)が0より大きいか小さいかで拮抗筋主動筋が求まるとは限らない。
  つまり、これにもある程度誤差があるということである。
  C=0とすることでmuscle jacobianの誤差による拮抗が起こり得るという問題はあるが、間違った拮抗関係を作ってしまう筋はモーメントアームの小さいものだけなため、大きく内力が溜まるということはない。
  また、$C$を負方向に大きくすれば、動かしたい方向に少しだけしかモーメントアームのない主動筋がみな抑制され、拮抗筋抑制の効果が大きくなる。
  つまり、誤差があっても拮抗関係が起こらないようにすることが可能なのである。
  これによって、動作したくない方向への動きが多少許容されるため、うまく主動筋間の筋張力が釣り合うような方向に動作する現象が見られる。
  しかし、それは同時に意図しない動きを生み、位置の精度に関して問題が生じうる。
  逆に$C$を正方向に大きくすれば、動作を妨げる方向に少しだけしかモーメントアームのない拮抗筋が主動筋として参加し、拮抗筋抑制の効果は薄れるものの位置精度の問題は少なくなる。
  次に、関節角度推定値に関してである。
  複雑な関節の関節角度推定値はあまり正確とは言えない。
  しかし、拮抗筋抑制制御ではモデル通りに実機の主動筋を動作させ、主動筋拮抗筋の判定のみに正確さに劣る関節角度推定値を用いる。
  拮抗関係はモデル誤差が生じにくいため、関節角度推定値の誤差はあまり問題にならない。
}%

\switchlanguage%
{%
  \subsection{Basic Experiment of Elbow Joint}
  Before the wide range limb motion experiment, we conducted an easy experiment of the uniaxial elbow joint.
  The uniaxial elbow joint of Kengoro is composed of three muscles: the triceps brachii, brachialis, and biceps brachii.
  We used a simple model representing the route of muscles by only the starting and end points as the geometric model of the elbow joint for easy understanding of the result.
  We moved the elbow joint of Kengoro up to 90 [deg] by 30 [deg] using joint-based AIC (JAIC), muscle-based AIC (MAIC), muscle stiffness control (MSC), and joint space control (JSC, \cite{humanoids2016:kawamura:controll}).
  We show the result in \figref{figure:basic-experiment}.

  \begin{figure}[htb]
    \centering
    \includegraphics[width=1.0\columnwidth]{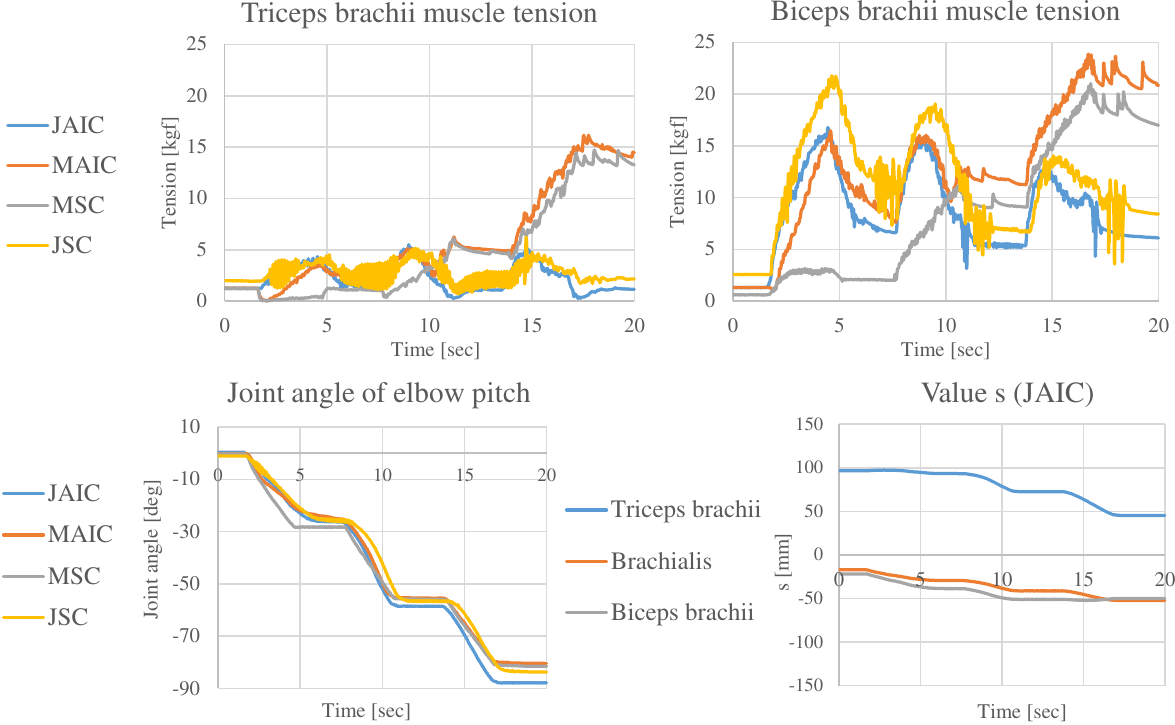}
    \vspace{-3.0ex}
    \caption{Basic experiment of elbow joint. Upper graph shows comparison of muscle tensions of biceps and triceps brachii among joint-based AIC (JAIC), muscle-based AIC (MAIC), muscle stiffness control (MSC), and joint space control (JSC). Lower left graph shows joint angle of the elbow pitch, and lower right shows value $s$ of JAIC.}
    \label{figure:basic-experiment}
  \end{figure}

  We can see similar behavior regarding internal muscle tension between MAIC and MSC.
  This is because the antagonistic state of MAIC finally comes to state 13 by model error as mentioned above, and agonist and antagonist muscles attract strongly together.
  Compared with JAIC, the joint angle trajectories of MAIC and MSC are poor at -90 [deg], but MSC is stable while the movement is small because of cocontraction.
  JAIC and JSC also show similar behavior, but JSC causes muscles to vibrate easily and is poor at joint angle trajectory because of a torque control using only muscle jacobian.
  JSC is influenced strongly by the model error of muscle jacobian; for example, Kengoro can raise the complex shoulder roll joint by only 60 [deg] in response to a command of 120 [deg].
  On the other hand, in JAIC, from the value $s$, we can see that brachialis and biceps brachii are always agonist muscles and triceps brachii is always an antagonist muscle, so large internal muscle tension is avoided.
  Also, even if $\bm{\theta}$ surpasses $\bm{\theta}_{target}$, the positive and negative of this value $s$ merely becomes reversed, and large internal muscle tension can be avoided completely.

  For these reasons stated in the subsections, AIC can move the actual robot without fatal damage by using a somewhat accurate geometric model of the complex tendon-driven musculoskeletal humanoid, which is difficult to model.
  In addition, AIC can fulfill its significance as the human simulator because it is based on the human nervous system.
}%
{%

  \subsection{Basic Experiment of Elbow Joint}
  広可動域動作を行う前に、肘の一軸を用いて簡単な実験を行った。
  腱悟郎の肘の一軸は上腕三頭筋・上腕筋・上腕二頭筋の3つの筋で構成されている。
  肘一軸のモデルとしては、実験結果がわかりやすいように筋肉の起始点と終始点のみで表された簡単なモデルを使用している。
  肘を0度から-90度まで、-30度ずつ3秒進み3秒休みを順々に行い、joint-based AIC, muscle-based AIC, MSC, Joint Space Controlでその際の比較を行った。
  その際の結果を\figref{figure:basic-experiment}に示す。
  MAICとMSCの内力の高まりは同じような挙動を示している。
  これは、MAICが上に述べられたようにモデルと実機の誤差によって最終的に状態13に落ち着き、最終的に拮抗筋と主動筋が大きく引き合ってしまっているからである。
  JAICと比べるとMAIC・MSCは内力の高まりによって-90度での追従は悪いが、MSCは常に筋が引き合っているため、動作角が小さく内力が小さい状況では非常に安定していると言える。
  また、JAICとJSCは似た挙動を示しているが、JSCは筋長ヤコビアンのみを使ったトルク制御のため振動しやすく関節角度の追従も遅い。
  JSCは筋長ヤコビアンの誤差に大きな影響を受けるため、例えば複雑な肩関節のRoll方向動作では120度指令に対して60度程度しか腕が上がっていない。
  JAICでの$s$の値を見ると、上腕筋と上腕二頭筋は常に主動筋、上腕三頭筋は常に拮抗筋となっていることが読み取れ、内力の高まりを回避できていることがわかる。
  また、もし$\bm{\theta}$が$\bm{\theta}_{target}$を超えてしまったとしても、この値sの正負が全て逆になるだけであるため、必ず内力の高まりを回避できる。

  これらアプローチは複雑でモデル化の難しい筋骨格腱駆動ヒューマノイドをある程度の正確さの幾何モデルでも実機破損を回避しつつ動作させることができる。
  また、拮抗筋抑制制御は人間にも取り入れられているため腱悟郎の人体シミュレータとしての意義も全うできる制御であると考える。
}%

\section{Achievement of Wide Range Upper Limb Motion by Antagonist Inhibition Control} \label{sec:5}
\switchlanguage%
{%
  We conducted experiments on wide range upper limb motion, which can be safely achieved using AIC.
  First, we will show the effectiveness of this control based on experiments conducted on the shoulder, and then on the scapula.
  Second, we conducted experiments of raising the arm using the shoulder and scapula, and checked the feasibility of this motion.
  For comparison, we used the basic control method, muscle stiffness control (MSC), in which $K_{stiffness}$ is constant.
  Finally, we conducted dangling and pull-up experiments, which become possible as a result of avoiding the internal muscle tension due to model error.
  In all of these experiments using AIC and MSC, $T_{bias}$ is 2 [kgf], $k$ is 10, $C$ is 0, and $t_k$ is 1000 [msec].
}%
{%
  拮抗筋抑制制御によって安全に実現することのできた広可動域動作について実験を行う。
  まず、肩単体、肩甲骨単体による拮抗筋抑制制御の動作テストを行い、この制御の有用性を示す。
  次に、肩甲骨と肩を使った、手を真上まで挙げる広可動域動作を行い、その動作実現性を確認する。
  比較対象として、本論文で開発した拮抗筋抑制制御の$K_{stiffness}$が一定である単純な筋剛性制御を行うこととする。
  最後に、モデル誤差による内力の高まりを回避した結果可能となった長時間のぶら下がり・懸垂動作を行い、本研究の締めくくりとする。
  この際の拮抗筋抑制制御・筋剛性制御においてはすべて、$T_{bias}$は2[kgf], $k$は10, $C$は0として実験を行った。
}%

\begin{figure}[htb]
  \centering
  \includegraphics[width=1.0\columnwidth]{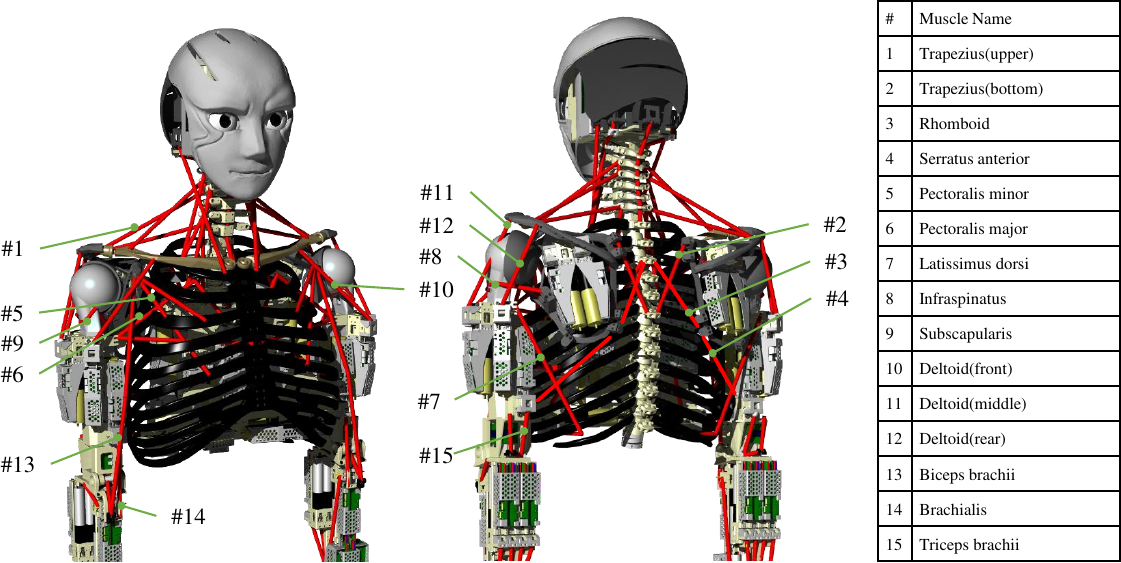}
  \vspace{-3.0ex}
  \caption{Muscle arrangement of Kengoro upper limb. $\#1$--$\#7$ is related to motion of the scapula and $\#6$--$\#15$ is related to motion of the shoulder.}
  \label{figure:kengoro-upper-limb-muscle}
\end{figure}

\subsection{Movement Using the Shoulder by AIC}
\begin{figure}[htb]
  \centering
  \includegraphics[width=1.0\columnwidth]{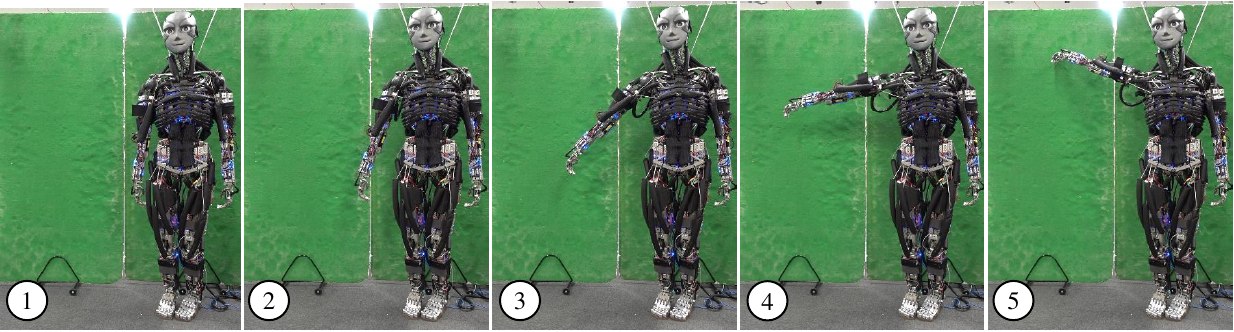}
  \vspace{-3.0ex}
  \caption{Experimental motion of shoulder.}
  \label{figure:only-shoulder-motion}
\end{figure}

\begin{figure}[htb]
  \centering
  \includegraphics[width=1.0\columnwidth]{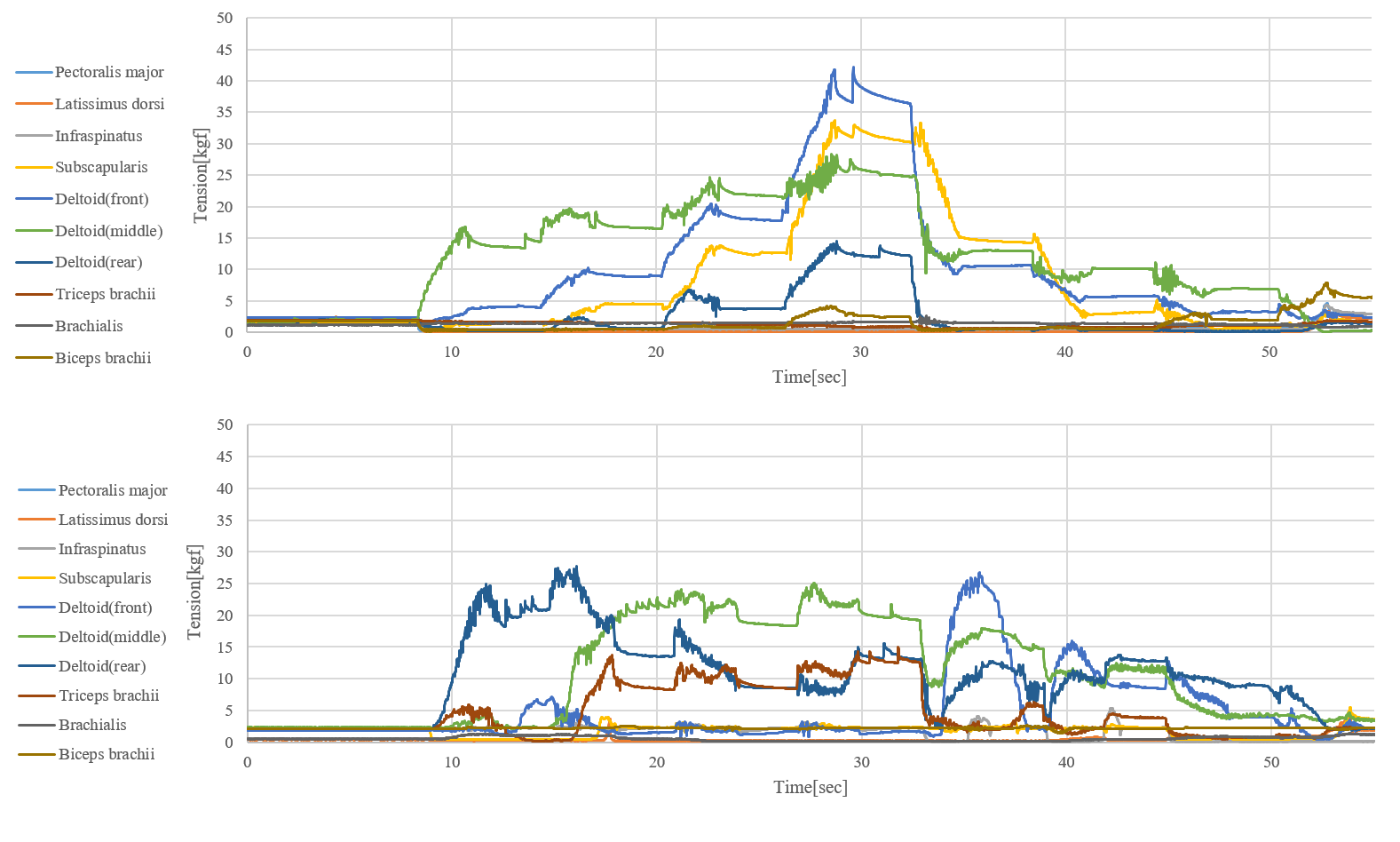}
  \vspace{-3.0ex}
  \caption{Result of shoulder motion experiment. Comparison of muscle tension between muscle stiffness control (upper graph) and antagonist inhibition control (lower graph).}
  \label{figure:only-shoulder-log}
\end{figure}

\switchlanguage%
{%
  The shoulder is driven by ten muscles, $\#6$--$\#15$, of those in \figref{figure:kengoro-upper-limb-muscle}.
  The experimental motion is doing abduction as in \figref{figure:only-shoulder-motion}, and doing adduction after that.
  The results of this experiment are indicated in \figref{figure:only-shoulder-log}.
  In this experiment, the maximum muscle tension decreased from 43 [kgf] to 28 [kgf] in AIC as compared to MSC.
  Especially, the subscapularis muscle and deltoid(front) display very high muscle tension in MSC, but this muscle tension is not displayed at all in AIC.
  This is because the subscapularis muscle does not have moment arm in the direction of abduction, but has moment arm in the direction of external rotation, and so the muscle is inhibited completely in AIC.
  The subscapularis muscle exhibits large model error, and thus muscle tension emerges in the unintended direction of external rotation.
  To keep the current joint angle against this motion, the deltoid(front) must exhibit large muscle tension, and this causes wasteful muscle tension by pulling at each other unnecessarily in MSC.
  Additionally, from 40 [sec], three deltoid muscles have even muscle tension because muscles with small moment arm to move in the intended direction are inhibited, movements other than abduction and adduction are permitted, and small muscle load sharing occurs, as discussed in the previous section.
  An example of a study in muscle load sharing is \cite{humanoids2013:asano:loadsharing}.
  In this example, agonist muscles must be chosen ourselves, but we can inhibit antagonist muscles and share muscle load at the same time using the antagonistic decision part of this study.
}%
{%

  肩は\figref{figure:kengoro-upper-limb-muscle}のうちの$\#6$--$\#15$の10本の筋で駆動されている。
  実験動作としては\figref{figure:only-shoulder-motion}のようにを外転を行い、その後腕を降ろす動作である。
  実験結果は\figref{figure:only-shoulder-log}のようになっている。
  この場合、筋剛性制御よりも拮抗筋抑制制御のほうが最大筋張力が43[kgf]から28[kgf]にまで下がっている。
  特筆すべきは、筋剛性制御では肩甲下筋(Subscapularis), Deltoid(front)が非常に大きな張力を発揮しているのに対し、拮抗筋抑制制御では一切張力を発揮していないことである。
  これは、肩甲下筋が外転方向に対してモーメントが全くなく、外旋に対してモーメントのある筋であるため、拮抗筋抑制制御では完全に抑制されていることが理由である。
  肩甲下筋はモデル誤差が非常に大きく、それゆえ動作させたくない外旋方向に対して力を発生させてしまう。
  それを逆方向に補うためにDeltoid(front)は大きな力を出さなければならず、筋剛性制御では無駄に張力を発揮しあってしまっているのである。
  対して拮抗筋抑制制御では、動作させたい方向にモーメントがない筋は抑制されるため無駄な張力を発揮せず、全体として張力のピークが非常に小さくなっているのである。
  また、40[sec]〜ではDeltoidの3本の筋張力が均等に割り振られているが、これは動かしたい方向に対するモーメントの小さな筋が指令に追従せずに外転内転以外の自由度を多少許すことで、主動筋間の負荷が分散されている現象だと思われる。
  主動筋間の負荷分散の研究としては\cite{humanoids2013:asano:loadsharing}が例として挙げられるが、これは主動筋を人が自ら選ばねばならず、本研究の拮抗筋抑制制御の主動筋拮抗筋判断部を利用することにより、拮抗筋の抑制と主動筋間負荷分散が同時にできるであろうと考える。
}%


\subsection{Movement Using Scapula by AIC}
\begin{figure}[htb]
  \centering
  \includegraphics[width=1.0\columnwidth]{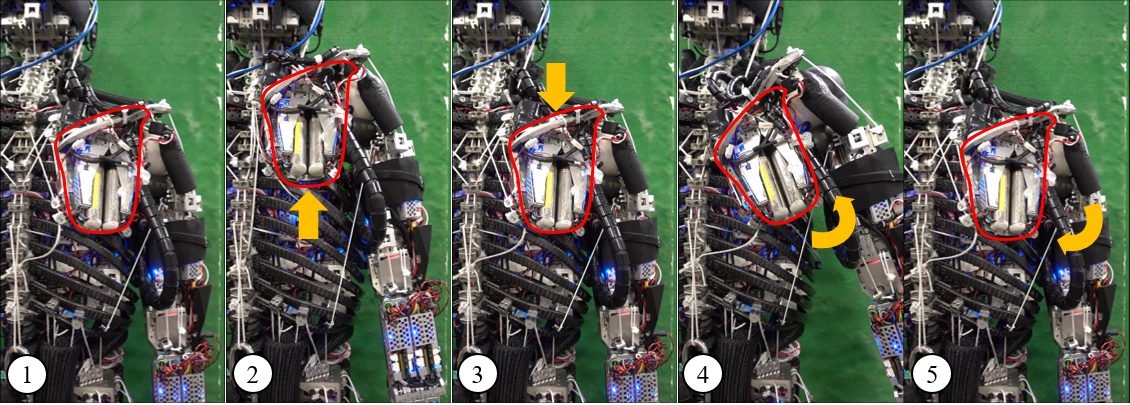}
  \vspace{-3.0ex}
  \caption{Experimental motion of scapula.}
  \label{figure:only-scapula-motion}
\end{figure}

\begin{figure}[htb]
  \centering
  \includegraphics[width=1.0\columnwidth]{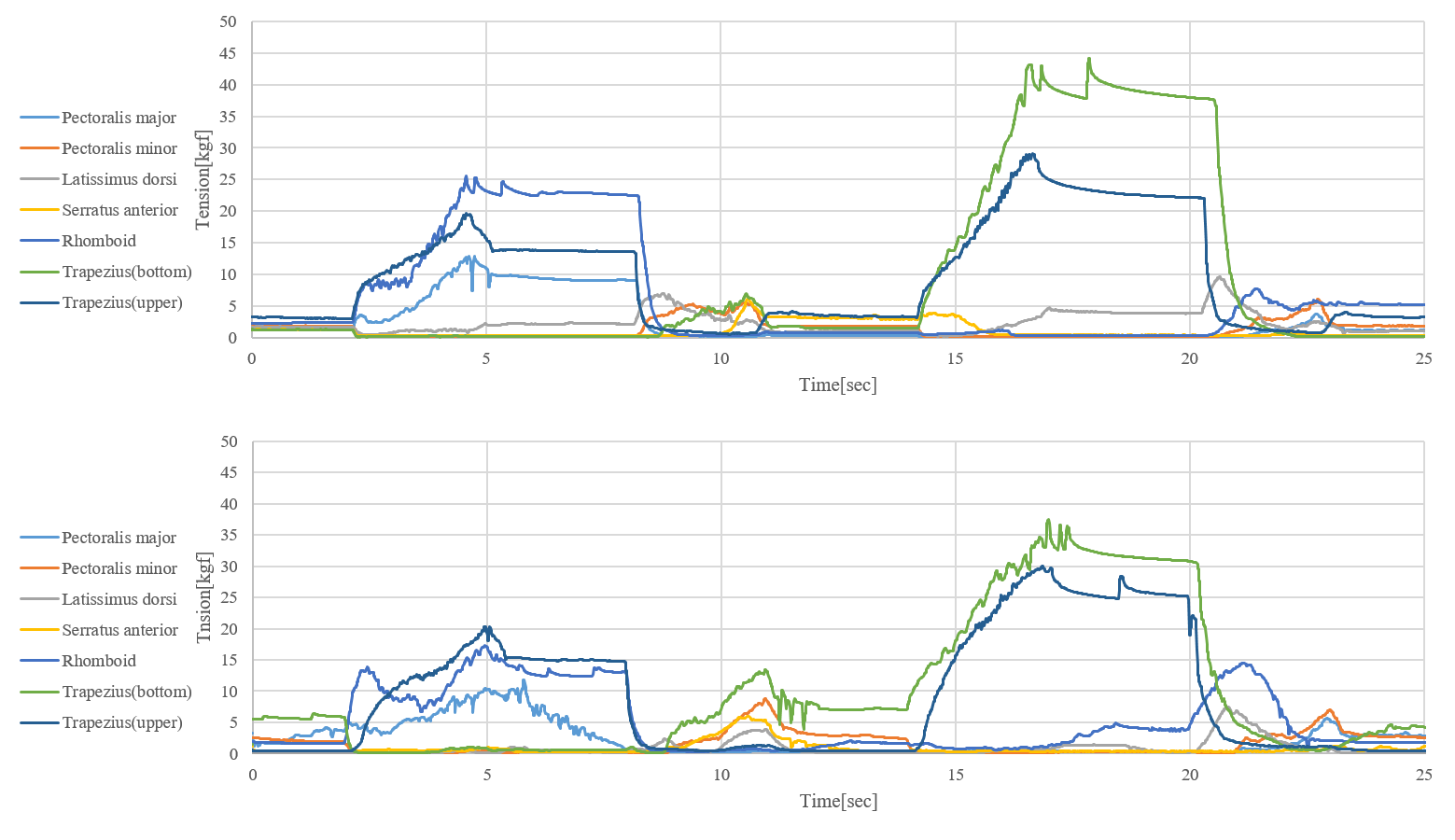}
  \vspace{-3.0ex}
  \caption{Result of scapula motion experiment. Comparison of muscle tension between muscle stiffness control (upper graph) and antagonist inhibition control (lower graph).}
  \label{figure:only-scapula-log}
\end{figure}

\switchlanguage%
{%
  The scapula is driven by seven muscles, $\#1$--$\#7$, of those in \figref{figure:kengoro-upper-limb-muscle}.
  The experimental motion is doing elevation, depression, upward rotation, and downward rotation continuously as shown in \figref{figure:only-scapula-motion}.
  The results of this experiment are indicated in \figref{figure:only-scapula-log}.
  In AIC, the antagonist muscle, the Latissimus dorsi, is inhibited and this decreases the peak of muscle tension.
  However, Kengoro's scapula has only a small amount of muscle per freedom of scapula, so antagonistic muscles hardly emerge.
  Additionally, the workspace of joints is small, so the model error is small as well.
  Thus, although the maximum muscle tension decreased from 44 [kgf] to 37 [kgf], AIC was not as effective as it was in movement using the shoulder.
}%
{%
  肩甲骨は\figref{figure:kengoro-upper-limb-muscle}のうちの$\#1$〜$\#7$の7本の筋で駆動されている。
  実験動作としては\figref{figure:only-scapula-motion}のように挙上下制・上方下方回旋を連続して行う。
  実験結果は\figref{figure:only-scapula-log}のようになっている。
  この場合、筋張力の時間変化に関してある程度の効果が得られている。
  拮抗筋抑制制御では、上方下方回旋において拮抗筋である広背筋(Latissimus dorsi)が抑制されており、筋張力のピークが減っている。
  しかし、肩甲骨関節は自由度に対して筋が非常に少なく拮抗関係が生まれにくい。
  また、可動域が狭いためモデル誤差が乗りにくい。
  そのため、筋張力最大値は44[kgf]から37[kgf]まで落ちてはいるものの、肩ほどの効果は得られていないように見える。
}%

\subsection{Movement Using Scapula and Shoulder by AIC}
\begin{figure}[htb]
  \centering
  \includegraphics[width=1.0\columnwidth]{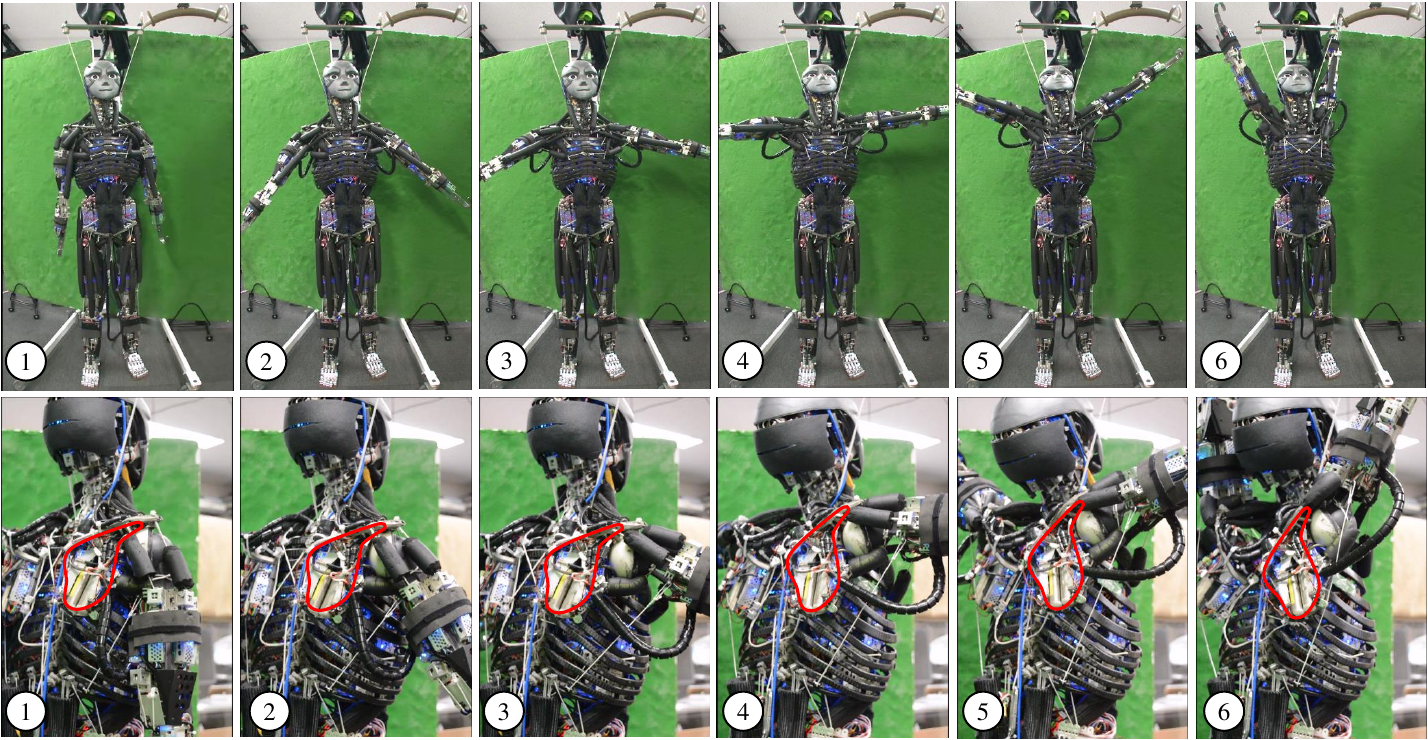}
  \vspace{-3.0ex}
  \caption{Experimental motion of raising arm by using scapula and shoulder. Front view (upper) and side view (lower).}
  \label{figure:scapula-shoulder-motion}
\end{figure}

\begin{figure}[htb]
  \centering
  \includegraphics[width=1.0\columnwidth]{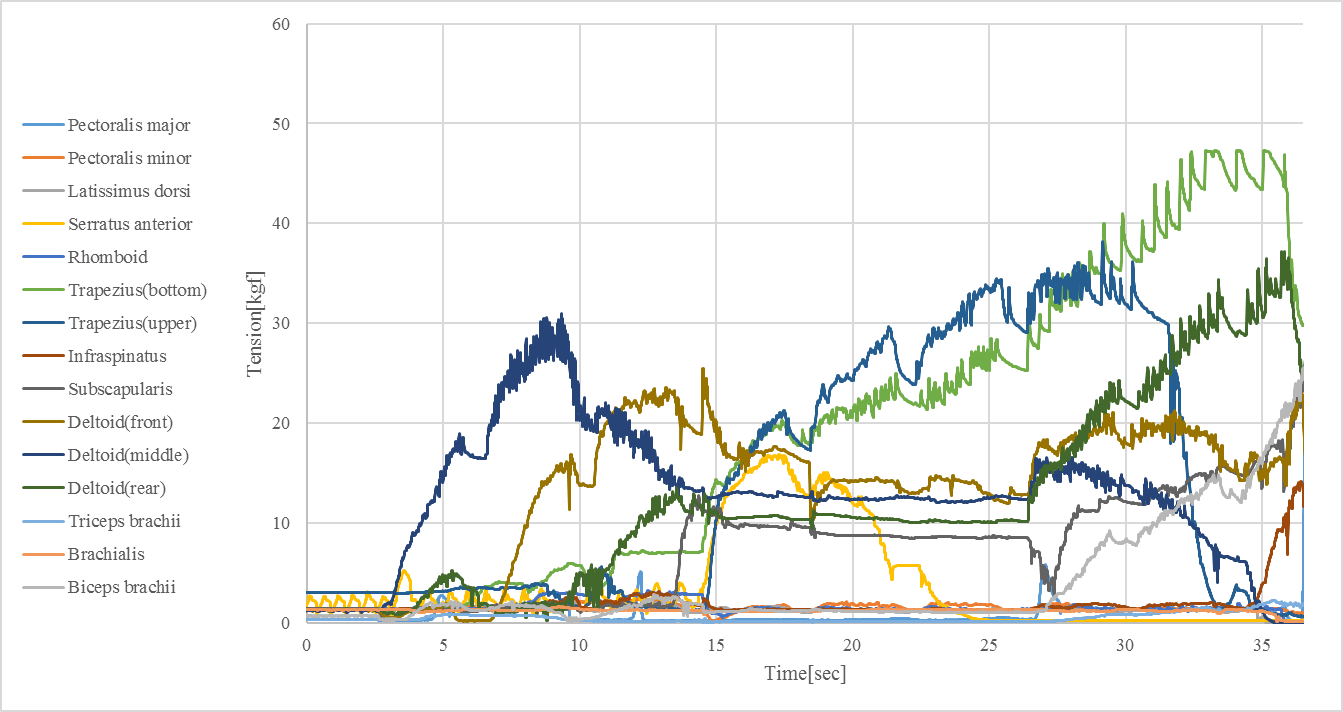}
  \vspace{-3.0ex}
  \caption{Result of raising arm motion experiment. Comparison of muscle tension between muscle stiffness control (upper graph) and antagonist inhibition control (lower graph).}
  \label{figure:scapula-shoulder-log}
\end{figure}

\switchlanguage%
{%
  We conducted wide range motion experiments of raising the arm upwards using the scapula and shoulder.
  In this movement, the scapula rotates upward by 60 [deg] when the abduction of the shoulder is by 120 [deg], and finally raises the arm to 180 [deg].
  The experimental movement is raising the arm and putting them down continuously as shown in \figref{figure:scapula-shoulder-motion}.
  The results of this experiment are indicated in \figref{figure:scapula-shoulder-log}.
  First, the maximum muscle tension is 55 [kgf] in MSC and is 45 [kgf] in AIC, showing that AIC can ease muscle tension.
  In MSC, the trapezius(upper), which shows maximum muscle tension, emerges beyond the tension limit of the tension measurement unit, leveling off the muscle tension, and is actually thought to be a higher value of about 60 to 70 [kgf].
  Second, regarding the distribution of high tension muscles, in AIC, the infraspinatus muscle and subscapularis muscle, which do not have moment arm in the direction of abduction, are inhibited, but in MSC, the two muscles exhibit high tension against each other, as in the experiment on the shoulder motion.
  Additionally, in AIC, the trapezius(upper and bottom) is used to move the scapula, but in MSC, only the trapezius(upper) is used due to model error, and very high tension emerges.
  Finally, muscle tension increases like that of a human in AIC.
  In MSC, maximum muscle tension emerged when raising the arm upward completely, but in the case of a human, we need maximum torque and muscle tension when raising the arm to 90 [deg].
  In AIC, model error is absorbed and only necessary muscle tension emerges, so muscle tension increases like that of a human.
}%
{%
  肩甲骨と肩を使った手を真上まで挙げる広可動域動作を行う。
  これは肩を120度外転する間に肩甲骨が一緒に60度上方回旋する動作であり、最終的に腕が180度まで回転する動作である。
  実験動作としては、\figref{figure:scapula-shoulder-motion}のように腕を真上まで上げ、その後下ろしていく。
  実験結果としては\figref{figure:scapula-shoulder-log}のようになっている。
  まず、最大筋張力としては筋剛性制御で55[kgf]、拮抗筋抑制制御で45[kgf]と10[kgf]程度拮抗筋抑制制御の方が筋張力を緩和できている。
  筋剛性制御で最大筋張力を発揮している僧帽筋上部に関しては張力測定ユニットの定格を超えており、値が飽和しているため、実際にはさらに大きな値となっていると考えられる。
  次に、大きな筋張力を発揮している筋の分布であるが、肩関節単体動作のときと同様に、拮抗筋抑制制御では直上拳上方向にほとんどモーメントアームのない棘下筋と肩甲下筋が抑制されているのに対して、筋剛性制御はその二つが大きな筋張力を発揮しあってしまっている。
  また、拮抗筋抑制制御では肩甲骨を主に僧帽筋下部と僧帽筋上部を用いて動作させているのに対して、筋剛性制御ではモデル誤差の関係から僧帽筋上部のみに動作を頼っており、大きな張力が出てしまっている。
  最後に、拮抗筋抑制制御は人間らしい筋張力の高まりをしていると考える。
  筋剛性制御ではモデル誤差によって腕を真上まで上げた際に最大筋張力を発揮してしまっているが、人間の場合、腕を直角に保つ状態が最も大きく重力補償トルクが必要であり、筋張力が高まると考える。
  拮抗筋抑制制御では実機とモデルの誤差を吸収することができ、その動作に必要な筋のみに力が働いているため、人間に近い筋張力の高まりとなっているのである。
}%

\subsection{Achievement of Long Time Dangling and Pull-up}
\begin{figure}[htb]
  \centering
  \includegraphics[width=1.0\columnwidth]{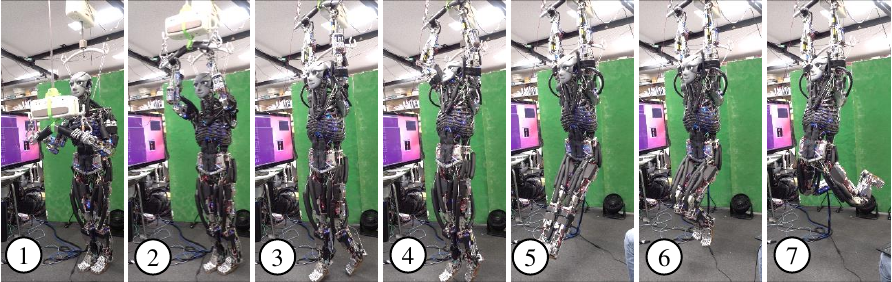}
  \vspace{-3.0ex}
  \caption{Motion of pull-up (3--4) and dangling (5--7).}
  \label{figure:kensui-motion}
\end{figure}

\begin{figure}[htb]
  \centering
  \includegraphics[width=0.95\columnwidth]{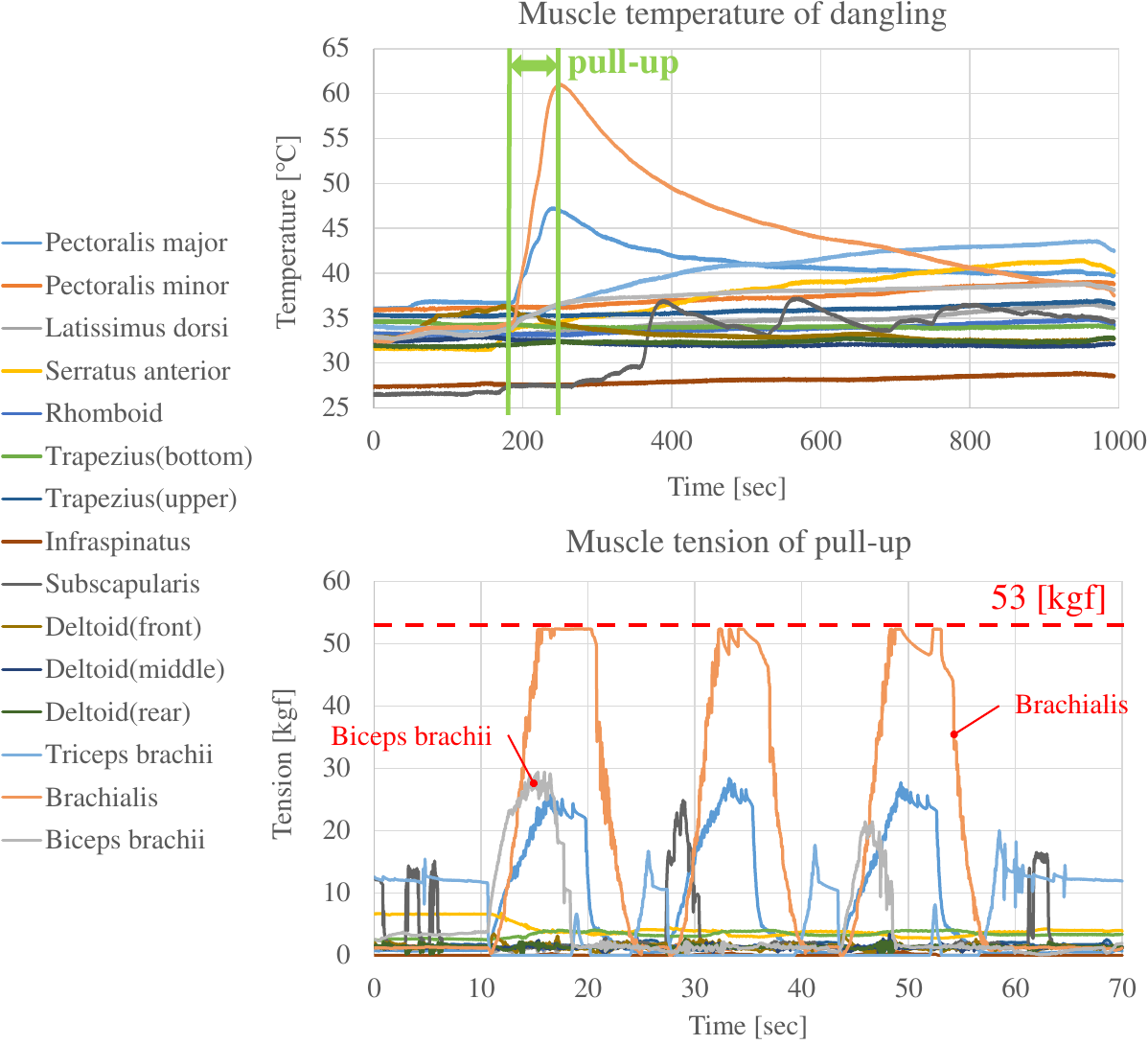}
  \caption{Muscle temperature of dangling (upper graph) and muscle tension of pull-up motion (lower graph).}
  \label{figure:kensui-logs}
\end{figure}

\switchlanguage%
{%
  As presented so far, the simple control method of MSC, which follows target muscle length, is very dangerous because internal muscle tension due to model error increases as wide range motion becomes wider.
  Also, robots can only continue wide range motion for a short time due to the increase of temperature.
  As a final experiment of this study, we performed the dangling motion for a long time in order to verify the effectiveness of AIC.
  At the same time, we performed the pull-up motion, and considered future tasks based on the increase of muscle tension.
  As a reference, in Kengoro, links in the upper body cannot support the body by themselves like in ordinary robots, because the shoulder in a tendon-driven musculoskeletal humanoid is a ball joint and can dislocate.
  The motion results of the experiment are shown in \figref{figure:kensui-motion}.
  Kengoro holds onto a rod, lifts the rod while sending target length to raise the arms, and dangles.
  We did the pull-up motion three times, and made various poses while dangling for 14 minutes after that.
  As shown in upper graph of \figref{figure:kensui-logs}, except for the pull-up motion, the muscles did not increase in temperature, and we were able to show that robots can realize wide range motion safely for a long time using AIC.
  Additionally, the muscle tension during a pull-up is shown in lower graph of \figref{figure:kensui-logs}.
  We were able to avoid high internal muscle tension by using AIC, and the subscapularis muscle, which normally has large model error, did not exhibit high tension at all.
  However, Kengoro could not move the elbow joint correctly during the pull-up motion.
  This is thought to be because the brachialis muscle had achieved maximum tension.
  To move the elbow joint correctly, Kengoro needs the cooperation of the brachialis muscle and biceps brachii muscle, and integration of muscle load sharing \cite{humanoids2013:asano:loadsharing} with AIC.
}%
{%
  これまでに見たように、単純に筋長を追従させるような筋剛性制御では広可動域になるほど筋張力の高まりが生じ、非常に危険である。
  また、温度上昇により短時間しか広可動域動作を続けることができない。
  そこで本研究の締めくくりとして、長時間のぶら下がり動作を行い、拮抗筋抑制制御の効果を検証する。
  また、その際に懸垂動作を行い、筋張力の高まり方を見ることで今後必要な制御を考える。
  参考として、筋骨格腱駆動ヒューマノイドの肩は脱臼する球関節でありリンクとして繋がっていないため、通常のロボットのように力を加えなくてもリンク自体が体を支えてくれるわけではないことに留意されたい。
  実験動作結果としては、\figref{figure:kensui-motion}のようになっている。
  まず棒に掴まり、腕を上げるような指令を送りつつ棒を持ち上げ、ぶら下がる。
  懸垂動作を3回行い、その後様々なポーズをさせながら14分間ぶら下がった。
  その際、懸垂動作以外では筋に発熱はほとんどなく、拮抗筋抑制制御により安全に長時間の広可動域動作が実現できることが示された。
  また、懸垂動作の際の筋張力は\figref{figure:kensui-log}のようになった。
  拮抗筋抑制制御を用いることで筋張力の高まりを回避することができ、誤差の大きなねじり筋などの筋張力の高まりがほとんど発生していない。
  しかし懸垂であまり肘が曲がっていなく、これは上腕筋の筋張力が限界に達しているからだと考える。
  さらに肘のトルクを増やすためには、上腕筋と上腕二頭筋の協調が必要であり、今後、主動筋間の負荷分散機構\cite{humanoids2013:asano:loadsharing}などと拮抗筋抑制制御を統合する必要があると考える。
}%

\section{CONCLUSION} \label{sec:6}
\switchlanguage%
{%
  In this study, we explained the realization of wide range motion using antagonist inhibition control (AIC), which is based on reciprocal innervation in the human nervous system.
  By deciding between agonist or antagonist muscle from the muscle jacobian and inhibiting antagonist muscles, we can prevent inhibition of motion due to model error, fatal damage to the muscle and structure, and muscle slack without a perfect geometric model.
  Through the experiments of the shoulder and scapula, we showed that AIC can solve the problem that the wider the motion is, the more the internal muscle tension increases in the usual method of movement called Muscle Stiffness Control (MSC).
  We also showed that wide range motion such as pull-up and dangling, which is very dangerous due to the accumulating internal muscle tension, can be done safely for a long time by using AIC.

  In future works, since wide range motion was achieved, we aim to do these motions more speedily and accurately through machine learning of model error.
  Because the tendon-driven musculoskeletal humanoid has a complex body based on a human, we believe that the meaning of this study lies in the acquisition of the self-body beyond achieving wide range motion.
}%
{%
  本論文では、人体の反射である相反性神経支配を模した拮抗筋抑制制御による動作範囲の拡張について述べた。
  完全に正確な幾何モデルでなくとも、筋張ヤコビアンと関節角度推定値から主動筋か拮抗筋かを判定し拮抗筋を抑制することによって、幾何モデルと実機の誤差による動作の妨げや、筋や骨格の破損、筋の弛みを防ぐことができることを提案した。%
  拮抗筋抑制制御の肩甲骨や肩への適用を通して、通常の筋剛性制御では広可動域になればなるほど内力の溜まる状態を解消できることを示した。%
  そして、今まで大きく内力が溜まるために非常に危険だったぶら下がりや懸垂という広可動域動作を、拮抗筋抑制制御により安全に長時間実現ができることを示した。%

  今後としては、広可動域を実現することができたため、その誤差の分を学習することでより広可動域動作を素早く、正確に行えることを目指す。
  複雑な身体を有する人体模倣ヒューマノイドを用いているため、この研究の意義は広可動域動作の先の自己身体の獲得ということに本当の意味があると考えるからである。
}%

{
  \bibliographystyle{IEEEtran}
  \bibliography{main}

\begin{thebibliography}{10}
\providecommand{\url}[1]{#1}
\csname url@rmstyle\endcsname
\providecommand{\newblock}{\relax}
\providecommand{\bibinfo}[2]{#2}
\providecommand\BIBentrySTDinterwordspacing{\spaceskip=0pt\relax}
\providecommand\BIBentryALTinterwordstretchfactor{4}
\providecommand\BIBentryALTinterwordspacing{\spaceskip=\fontdimen2\font plus
\BIBentryALTinterwordstretchfactor\fontdimen3\font minus
  \fontdimen4\font\relax}
\providecommand\BIBforeignlanguage[2]{{%
\expandafter\ifx\csname l@#1\endcsname\relax
\typeout{** WARNING: IEEEtran.bst: No hyphenation pattern has been}%
\typeout{** loaded for the language `#1'. Using the pattern for}%
\typeout{** the default language instead.}%
\else
\language=\csname l@#1\endcsname
\fi
#2}}

\bibitem{humanoids2013:michael:anthrob}
M.~J{\"a}ntsch, S.~Wittmeier, K.~Dalamagkidis, A.~Panos, F.~Volkart, and
  A.~Knoll, ``{Anthrob - A Printed Anthropomimetic Robot},'' in
  \emph{Proceedings of the 2013 IEEE-RAS International Conference on Humanoid
  Robots}, 2013, pp. 342--347.

\bibitem{humanoids2016:asano:kengoro}
Y.~Asano, T.~Kozuki, S.~Ookubo, M.~Kawamura, S.~Nakashima, T.~Katayama,
  Y.~Iori, H.~Toshinori, K.~Kawaharazuka, S.~Makino, Y.~Kakiuchi, K.~Okada, and
  M.~Inaba, ``{Human Mimetic Musculoskeletal Humanoid Kengoro toward Real World
  Physically Interactive Actions},'' in \emph{Proceedings of the 2016 IEEE-RAS
  International Conference on Humanoid Robots}, 2016, pp. 876--883.

\bibitem{robio2011:shirai:control}
T.~Shirai, J.~Urata, Y.~Nakanishi, K.~Okada, and M.~Inaba, ``{Whole body
  adapting behavior with muscle level stiffness control of tendon-driven
  multijoint robot},'' in \emph{Proceedings of the 2011 IEEE International
  Conference on Robotics and Biomimetics}, 2011, pp. 2229--2234.

\bibitem{csm1990:jacobsen:control}
S.~C. Jacobsen, H.~KO, E.~K. Iversen, and C.~C. Davis, ``{Control Strategies
  for Tendon-Driven Manipulators},'' in \emph{IEEE Control Systems Magazine,
  10}, 1990, pp. 23--28.

\bibitem{ijars:potkonjak:puller-follower}
V.~Potkonjak, B.~Svetozarevic, K.~Jovanovic, and O.~Holland, ``{The
  Puller-Follower Control of Compliant and Noncompliant Antagonistic Tendon
  Drives in Robotic Systems},'' in \emph{International Journal of Advanced
  Robotic Systems}, 2011, pp. 143--155.

\bibitem{robio2011:michael:control}
M.~J{\"a}ntsch, C.~Schmaler, S.~Wittmeier, K.~Dalamagkidis, and A.~Knoll, ``{A
  scalable Joint-Space Controller for Musculoskeletal Robots with Spherical
  Joints},'' in \emph{Proceedings of the 2011 IEEE International Conference on
  Robotics and Biomimetics}, 2011, pp. 2211--2216.

\bibitem{humanoids2016:kawamura:controll}
M.~Kawamura, S.~Ookubo, Y.~Asano, T.~Kozuki, K.~Okada, and M.~Inaba, ``A
  joint-space controller based on redundant muscle tension for multiple dof
  joints in musculoskeletal humanoids,'' in \emph{Proceedings of the 2016
  IEEE-RAS International Conference on Humanoid Robots}, 2016, pp. 814--819.

\bibitem{humanoids2013:asano:loadsharing}
Y.~Asano, T.~Shirai, T.~Kozuki, Y.~Motegi, Y.~Nakanishi, K.~Okada, and
  M.~Inaba, ``{Motion Generation of Redundant Musculoskeletal Humanoid Based on
  Robot-Model Error Compensation by Muscle Load Sharing and Interactive Control
  Device},'' in \emph{Proceedings of the 2013 IEEE-RAS International Conference
  on Humanoid Robots}, 2013, pp. 336--341.

\bibitem{humanoids2015:okubo:muscle-learning}
S.~Ookubo, Y.~Asano, T.~Kozuki, T.~Shirai, K.~Okada, and M.~Inaba, ``Learning
  nonlinear muscle-joint state mapping toward geometric model-free tendon
  driven musculoskeletal robots,'' in \emph{Proceedings of the 2015 IEEE-RAS
  International Conference on Humanoid Robots}, 2015, pp. 765--770.

\bibitem{iros2015:asano:module}
Y.~Asano, T.~Kozuki, S.~Ookubo, K.~Kawasaki, T.~Shirai, K.~Kimura, K.~Okada,
  and M.~Inaba, ``{A Sensor-driver Integrated Muscle Module with High-tension
  Measurability and Flexibility for Tendon-driven Robots},'' in
  \emph{Proceedings of the 2015 IEEE/RSJ International Conference on
  Intelligent Robots and Systems}, 2015, pp. 5960--5965.

\end{thebibliography}
}

\end{document}